\documentclass[pdflatex,sn-mathphys-num]{sn-jnl}

\usepackage{graphicx}%
\usepackage{multirow}%
\usepackage{amsmath,amssymb,amsfonts}%
\usepackage{amsthm}%
\usepackage{mathrsfs}%
\usepackage[title]{appendix}%
\usepackage[dvipsnames]{xcolor}%
\usepackage{float}
\usepackage{subcaption}
\usepackage{caption}
\usepackage{textcomp}%
\usepackage{manyfoot}%
\usepackage{booktabs}%
\usepackage{algorithm}%
\usepackage{algorithmicx}%
\usepackage{algpseudocode}%
\usepackage{listings}
\usepackage{bbm}
\usepackage{hyperref}
\theoremstyle{thmstyleone}%
\theoremstyle{thmstyletwo}%

\theoremstyle{thmstylethree}%

\begin{document}

\title[Article Title]{Hoeffding adaptive splitting trees for data stream classification with concept drift and ensemble learning}


    \author[1,2]{\fnm{Daniel} \sur{Nowak Assis}}\email{daniel.nassis@ppgia.pucpr.br}
    
    \author[1]{\fnm{Jean Paul} \sur{Barddal}}\email{jean.barddal@ppgia.pucpr.br}

    \author[1]{\fnm{Fabrício} \sur{Enembreck}}\email{fabricio@ppgia.pucpr.br}

    \affil[1]{\orgdiv{Programa de Pós-Graduação em Informática (PPGIa)}, \orgname{Pontifícia Universidade Católica do Paraná (PUCPR)}, \orgaddress{\street{Rua Imaculada Conceição 1155}, \postcode{80215-901}, \city{Curitiba}, \country{Brazil}}
    }
    \affil[2]{\orgname{Sorbonne Université, CNRS, LIP6},  \city{Paris}, \country{France}}



\abstract{Ensembles of decision trees are well-established methods for data stream classification. 
In ensemble learning, Hoeffding Trees are widely adopted as base learners, performing periodic split attempts according to the Hoeffding bound.
Recent studies, however, indicate that this standard splitting mechanism lacks adaptability, while adaptive trees that trigger splits in response to performance degradation have achieved superior results. 
In this paper, we identify limitations in the use of adaptive-splitting decision trees as ensemble base learners, showing that change detectors often fail to promote sufficient diversity within ensembles. 
To address this issue, we propose two novel decision tree models, termed Hoeffding Adaptive Splitting Trees. 
These models combine the periodic splitting strategy of Hoeffding Trees, which fosters ensemble diversity, with adaptive splitting mechanisms that employ change detection algorithms to identify performance decay and determine split points. 
Experimental results demonstrate that Hoeffding Adaptive Splitting Trees enhance ensemble performance and achieve state-of-the-art results across a comprehensive evaluation, including benchmark comparisons, computational cost analysis, and concept drift adaptation.
}



\maketitle

\section{Introduction}



Learning systems designed for online scenarios are well-known and an established area in Machine Learning. 
Data stream mining contemplates knowledge discovery in high-speed arriving data. 
In practice, it requires online deployment due to the high amount of data generated in real time. 
For an efficient learning system to work on streaming data, the algorithms deployed must be aware of the hasty and potentially infinite nature of streaming data. 
As pointed out in \cite{bifet2010moa}, ideally, a system designed for mining streaming data must \textit{i)} process an instance of data, analyze it one time, and discard it, \textit{ii)} use a limited amount of memory, \textit{iii)} process an instance as fast as possible, and \textit{iv)} be able to make predictions at any time.

Another challenge in streaming data is concept drift \cite{concept_drift_lu}. 
A concept drift occurs when the probabilistic properties of the data change over time. 
In a classification problem, a concept drift occurs between the times $t$ and $t + \Delta$, if $P_{t}(X, Y) \neq P_{t+\Delta}(X, Y)$, where $P_{t}$ refers to the joint distribution at time $t$ given a set of examples $X$ and their class labels $Y$. 
Concept drifts may decrease the accuracy of machine learning models, which must swiftly detect and adapt to those situations to avoid compromising the entire learning process.

Ensemble-based methods are considered state-of-the-art algorithms for the data stream classification problem. 
In the construction and evaluation of ensembles in the literature, the first choice as a starting point was, in general, the Hoeffding tree \cite{domingosHT2000}.
The Hoeffding Tree algorithm is an incremental method that gradually expands the tree by making periodic split attempts as the data streams evolve, relying on the principles outlined in the Hoeffding Theorem \cite{hoeffding1963probability}. 
It has long been considered a state-of-the-art approach for constructing decision trees in the context of mining streaming data. 
However, recent insights from \cite{last} highlight a limitation: the periodic split attempts may not effectively adapt to the evolving nature of the data stream. 
Since these split attempts occur at regular intervals, they fail to capture the precise moments when changes in accuracy or data distribution occur within the leaves. 
Furthermore, the algorithm continues to search for the best split even during periods of stability, causing increased processing time without significant improvements to the tree structure.
  
Local Adaptive Streaming Trees (LAST), introduced in \cite{last}, overcome this limitation by continuously monitoring tree performance or class-distribution purity and splitting whenever change detectors flag a change, thereby outperforming Hoeffding-based trees \footnote{
An ablation of LAST as a single tree has also been performed in \cite{kais_daniel}, where it depicts the behavior of LAST under different change detectors.
}. Despite these promising results, LAST lacks the strategies needed to serve as a base model for ensemble learners.
In practice, it has been observed that \textit{i)} change detectors operate per base model, so instances affect individual learners rather than the ensemble as a whole; \textit{ii)} the trees are learned without any diversity induction scheme; and \textit{iii)} updating change detectors repeatedly incurs high computational overhead.

Overall, it has been observed that \textit{i)} the high correlation among decision tree outputs in the initial moments of the stream affects change detection across base learners, inducing similar splitting times and low diversity; and \textit{ii)} splitting time can be affected by random sampling, and LAST's soft split constraint can produce poor splits, unlike the monolithic version, which reacts to changes in the model's true performance or distribution.
To address these challenges, we propose two new decision trees, namely Hoeffding Adaptive Splitting Trees (HASTs).
These models combine a periodic splitting mechanism, which induces diversity among the decision tree learners by producing different trees, with an adaptive splitting mechanism that tracks either leaf-node performance or class-distribution purity and can react to changes at the leaf-node level.
The two approaches differ in their combination strategy, drawing, respectively, on Hoeffding Trees \cite{domingosHT2000} and Extremely Fast Decision Trees \cite{efdt}.

The contributions of this paper are summarized as follows:

\begin{itemize}
    \item [1.] Two novel decision tree architectures that demonstrate state-of-the-art results in ensemble setups.
    \item [2.] Comprehensive experimental analysis with state-of-the-art ensembles and insights into their performance, computational cost, and how they react to concept drifts.
    \item [3.] Indication of future directions for the proposal of new ensembles.
\end{itemize}

The remainder of this paper is organized as follows. 
Section \ref{sec:background} details the fundamental concepts of data stream mining and concept drift. 
Section \ref{sec:lit_rev} discusses the algorithms used in this work, such as online decision trees and ensembles.
Section \ref{sec:method} revisits the limitations of LAST as base learners of ensembles and introduces the proposed decision trees. 
Section \ref{sec:met} outlines the experimental protocol used in this work. 
Section \ref{sec:res} discusses the results obtained. 
In Section \ref{sec:conc}, we conclude the paper by presenting the main findings of this study and identifying future research directions.

\section{Background and Definitions}
\label{sec:background}

Data streams are continuous and sequential data flows that arrive over time, with no predefined stopping point, often treated as potentially infinite. 
In a structured data stream, each arriving datum is a feature vector $x_t \in \mathbb{R}^d$. 
Classification (a prediction of a class $y_n$ given an input $\vec{x}_n$) in this framework fundamentally differs from batch classification because 
in streaming data, storing all data is unfeasible, and thus, efficient stream mining requires classifiers that can ideally process data instances as rapidly as they arrive. 
If processing lags, the only options are to ignore new data (rendering the classifier less current) or attempt to store it, risking memory exhaustion and system collapse.

Another problem in data streams is concept drift, characterized by temporal variations in the data's probability distribution. 
Specifically, drift occurs between time $t$ and $t+\Delta$ if the joint distribution $P(\vec{x},y)$ changes ($P_{t}(\vec{x},y) \neq P_{t + \Delta}(\vec{x},y)$). 
Based on the relationship $P(\vec{x},y) = P(\vec{x}) P(y|\vec{x})$, three types of drift are distinguished \cite{gama_survey_cd:2014}:
\begin{enumerate}
    \item \textbf{Virtual drift:} Changes solely in the marginal distribution of features, $P(\vec{x})$.
    \item \textbf{Real drift:} Changes solely in the posterior probability distribution (the classification boundary), $P(y|\vec{x})$.
    \item Changes affecting both $P(\vec{x})$ and $P(y|\vec{x})$ simultaneously \cite{concept_drift_lu}.
\end{enumerate}

In addition to the categories above, concept drifts can be categorized according to the temporal patterns exhibited. 
An abrupt shift is a sudden change from concept $C_1$ to $C_2$. 
A gradual shift is a slower transition between concepts. 
An incremental shift occurs through a series of intermediate concepts. 
Finally, a recurring concept scenario allows a previously observed concept to reappear.

For the interested reader in the concept drift problem, we recommend the following recent studies that broaden the topic \cite{concept_drift_lu, aguiar_drift}.

\section{Online Decision Trees and Ensembles}
\label{sec:lit_rev}

In this section, we introduce existing works on online decision trees and ensembles that use tree variants for data stream classification.

\subsection{Hoeffding Trees}
    Hoeffding trees \cite{domingosHT2000} are incremental and online decision trees that adopt a representation similar to their batch counterparts, such as C4.5 \cite{c4.5} and CART \cite{cart} algorithms.
    In contrast to standard decision tree learning, which performs a greedy evaluation of linear splits in a data batch, Hoeffding Trees continuously increment the statistics required to check whether a split should be performed, and this analysis is carried out periodically.

\begin{figure}[H]
    \centering
    \includegraphics[width=0.75\linewidth]{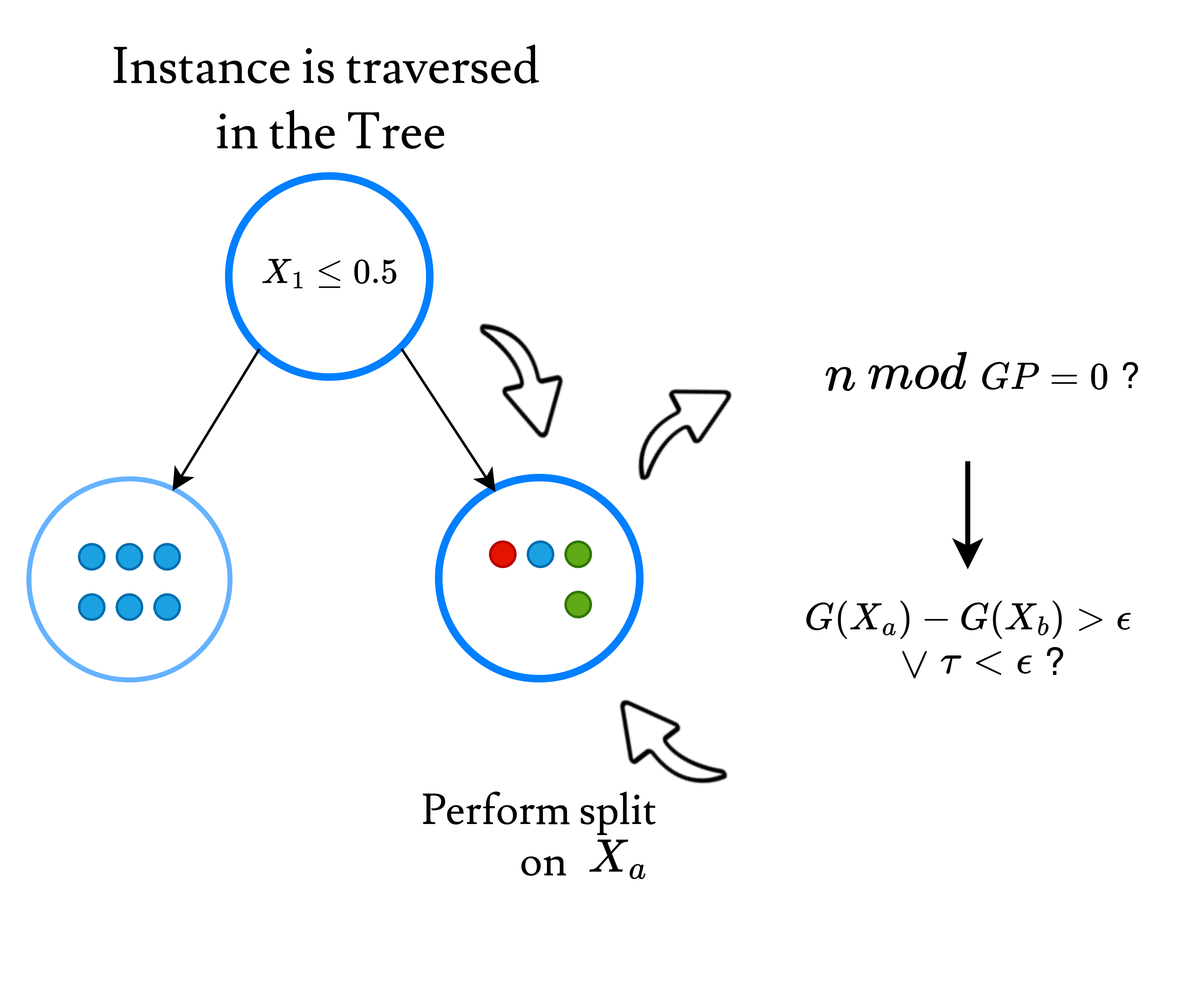}
    \caption{Training process of Hoeffding Trees}
    \label{tree:ht}
    \end{figure}

    Targeting efficiency, the two main components of Hoeffding Trees are the Hoeffding bound constraint to prevent overgrowth and the periodic evaluation of splits in leaf nodes. 
    Figure \ref{tree:ht} presents the incremental training and splitting mechanism of Hoeffding Trees, where the Grace Period ($GP$) is a user-given parameter that sets the frequency at which a leaf node will attempt to split. Denoting $n$ as the number of samples seen at leaf $l$, if $n$ mod $GP = 0$, a split attempt will ensue. Then it must pass the Hoeffding Test to result in a split.

    The Hoeffding bound \cite{hoeffding1963probability} is a theorem that presents a probability inequality for the difference between the mean ($\overline{X}$) and expected value ($\mathbbm{E}[X]$) of a set of random variables. 
    With a level of confidence $\delta$, one can derive that $\overline{X} - \mathbbm{E}[X] \geq \epsilon$ following Equation \ref{hoeff_ineq}, where $R$ is the range of the random variable, and $n$ is the number of observations.

    \begin{equation}
    \epsilon = \sqrt{\frac{R^{2}\log(\frac{1}{\delta})}{2n}}     \label{hoeff_ineq}
    \end{equation}
    
    The Hoeffding Tree applies this bound for impurity measures to determine if $X_{a}$, the attribute with the highest impurity measure $G$ (such as information gain~\cite{c4.5} or the Gini index~\cite{cart}), is the ideal attribute to split on a split attempt. Given $X_{b}$, the attribute with the second-highest impurity measure, if $G(X_{a}) - G(X_{b}) \geq \epsilon$, a split will occur on $X_{a}$.
    
    Note that $\lim_{n \to +\infty} \epsilon = 0$, and if $G(X_{a})$ and $G(X_{b})$ have similar values in a node, a split will require many observations to occur. To relax the Hoeffding constraint in these situations, the Hoeffding Tree has a tie threshold. Given $\tau$ (user-given threshold), a split will ensue when $\tau > \epsilon$.
    
    Assuming $d$ to be the number of attributes, $v$  the maximum number of values per attribute, $c$ the number of classes, and $l$ the number of leaf nodes, the Hoeffding tree algorithm requires $\mathcal{O}(ldvc)$ memory to store the necessary counts.

    An aspect that established Hoeffding Trees as accurate decision trees was the addition of Naive Bayes at leaf nodes \cite{nb_leaves_gama}. 
    It was observed that Naive Bayes could outperform decision trees in initial streaming scenarios where the decision tree has trained with little data. 
    In \cite{stress_testing}, the authors propose the selection of Naive Bayes or majority class strategy prediction according to the method with the highest accuracy at the leaf.

    Another critical aspect of Hoeffding Trees is memory management since storing all instances from a data stream is unfeasible. 
    Leaf nodes maintain statistics from data, such as histograms for nominal data, and rely on forgetting mechanisms. 
    We refer the interested reader to \cite{kirkbyPhdThesis} for more details.
    
\subsection{Extremely Fast Decision Trees}
\label{subsec:efdt}
    
    In \cite{efdt}, the authors propose EFDT (Extremely Fast Decision Trees), an extension of Hoeffding Trees. The first change to Hoeffding Trees refers to the Hoeffding Bound. 
    Instead of comparing the split evaluation function of the two features that maximize this function, the comparison is made between the feature $X_{a}$ that maximizes the split evaluation function and the case where no split happens ($X_{\emptyset}$). 
    Equation \ref{efdt_ineq} is used as a condition for performing a split in $X_{a}$.

\begin{equation}
    G(X_{a}) - G(X_{\emptyset}) > \sqrt{\frac{R^{2}\log(\frac{1}{\delta})}{2n}}     \label{efdt_ineq}
\end{equation}

Another extension proposed in EFDT is a split reevaluation mechanism \cite{efdt}. 
As more instances arrive, the probability of a better split appearing is higher than that of a split that ensued earlier. 
The reevaluation mechanism is similar to the splitting mechanism. 
The Hoeffding bound is used to compare the value of the split evaluation function of $X_{a}$ in case a new split replaces the tree branch by a new split and the current feature ($X_{current}$) split evaluation function at the time the split occurred. 
A branch of the tree is replaced by a new split if Equation \ref{efdt_reev} holds.

\begin{equation}
    G(X_{a}) - G(X_{\text{current}}) > \sqrt{\frac{R^{2}\log(\frac{1}{\delta})}{2n}}     \label{efdt_reev}
\end{equation}

Since EFDT must store data at intermediate nodes to re-evaluate splits, it requires $\mathcal{O}(ndvc)$ memory, where $n$ is the number of nodes. 
The default value for periodic reevaluation was defined as 2000 by the authors.

In \cite{efdt_ensemble}, the authors compare EFDT and Hoeffding Trees as base learners for ensembles, a similar aim to this work \cite{last}. 
Therefore, we also perform experiments with EFDT as a base learner of ensembles since the results in \cite{efdt_ensemble} suggest that EFDT is superior to Hoeffding Trees at an ensemble level. 
We also improve the evaluation of base learners in this work compared to \cite{efdt_ensemble} and explain the improvements in Section \ref{sec:met}.

\subsection{Adaptive Splitting Trees}

Hoeffding-based Trees follow a fixed periodic splitting strategy, activated at intervals of GP instances and controlled by the Hoeffding test.
This policy is invariant to stream dynamics and tree performance, leading to blind recurrent splitting attempts and greedy split searches when little change occurs in the leaf, which results in little change in the tree structure.

\begin{figure}[H]
    \centering
    \includegraphics[width=0.75\linewidth]{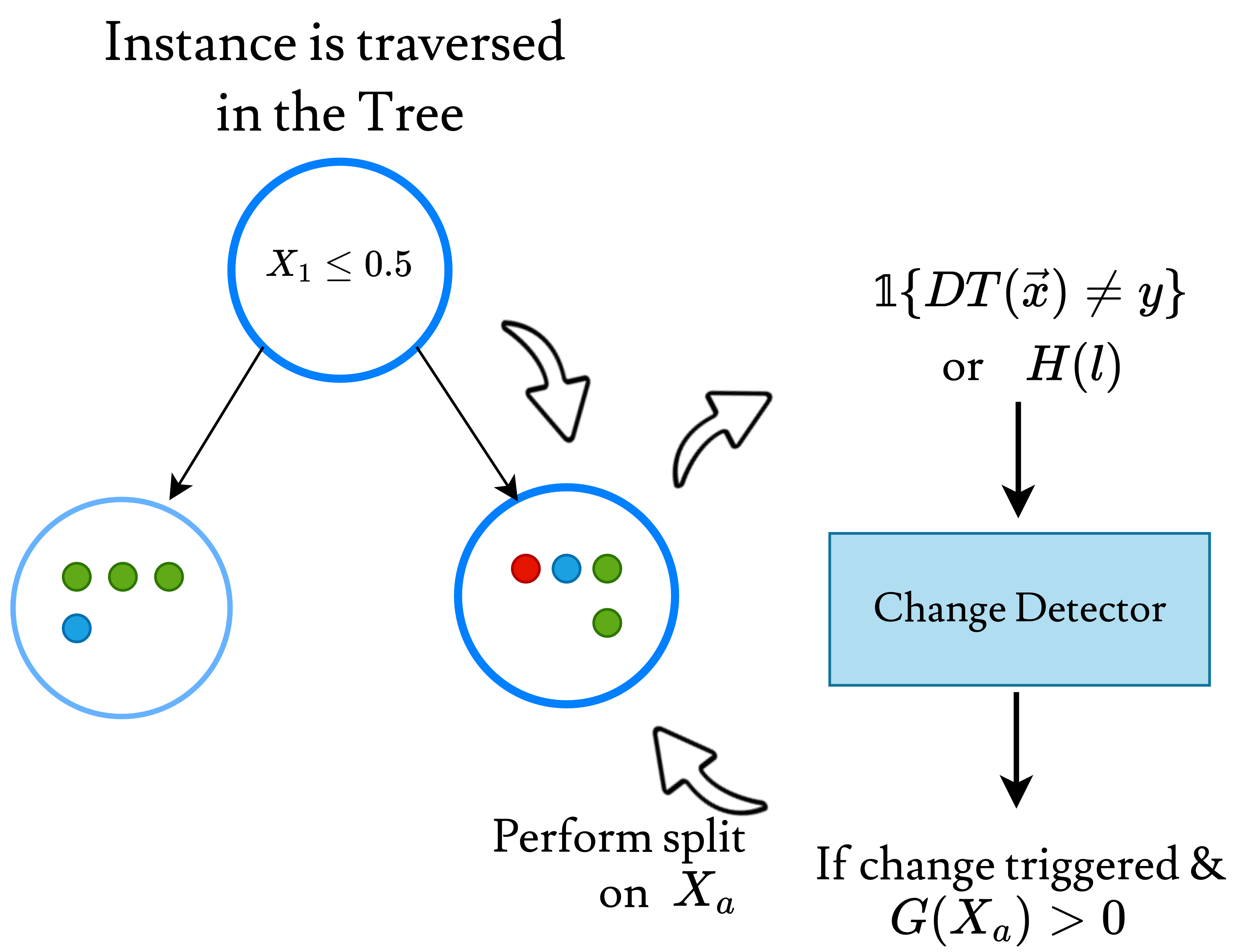}
    \caption{LAST's training process}
    \label{tree:last}
\end{figure}

In \cite{last}, the authors overcame this limitation by proposing the Local Adaptive Streaming Tree (LAST). 
The rationale behind LAST is that change detectors can provide adaptability to determine split points at leaf nodes. 
Figure \ref{tree:last} presents LAST's incremental training and splitting mechanism. 
In a leaf node, change detectors can monitor either the impurity or error rate. 
If a change detector at the leaf triggers a change, then a split will ensue if $G(X_{a}) > 0$.

Change detectors are constantly updated with arriving instances, meaning that these algorithms track how the stream evolves. 
Since LAST applies the softest split constraint ($G(X_{a}) > 0$), the change detectors effectively control how the tree grows.  

LAST requires $\mathcal{O}(\psi ldvc)$ memory, where $\psi$ is the memory complexity of the change detection algorithm applied.

In \cite{kais_daniel}, the authors presented a more thorough evaluation of LAST as a single monolithic decision tree, comparing LAST with different change detectors from the literature and performing an ablation analysis. This differs from the present paper, whose scope is ensemble learning.

\subsection{Online Ensembles}

    Ensembles are state-of-the-art algorithms for data stream classification, often adapting a batch version counterpart.   

    In \cite{oza_bagging}, the authors proposed an online version of Bagging \cite{bagging_breiman}. 
    For each incoming instance, a base learner trains with $k$ copies of the instance, where $k$ is a random variable that follows a Poisson$(\lambda = 1)$ distribution, simulating random sampling with
    replacement in online scenarios.
    This process induces diversity amongst its members, which have their votes cast and combined using simple majority voting.    
    In the same work, the authors proposed an online version of AdaBoost.M1 \cite{adaboost}. 
    For each classifier, correct and wrong classification weights are maintained. 
    When a new incoming instance arrives, the $\lambda$ parameter starts at one, and the classifier is trained with $k =$ Poisson($\lambda$) copies of the instance. 
    If a learner makes a wrong classification, the value $\lambda$ increases based on the learner's wrong weights, and in case the classifier makes a correct prediction, the value of $\lambda$ decreases for the training of the next classifier.
    In contrast to Online Bagging, Online Boosting assigns each ensemble member a weight, which is used during prediction in a weighted majority voting scheme.

    To deal with concept drifts, the authors in \cite{adwinbag} propose ADWIN-Bagging and ADWIN-Boosting. 
    For each ensemble classifier, an ADaptive WINdowing (ADWIN) \cite{adwin} drift detector monitors the error rates of the classifier. 
    If the detector coupled with a classifier detects a concept drift, the least accurate member of the ensemble (in terms of the estimated ADWIN error) is reset.

    Extending the ADWIN-Bagging algorithm, authors in \cite{lev_bag} propose the Leveraging Bagging algorithm. 
    Instead of sampling being conducted by Poisson($\lambda = 1$), sampling follows Poisson($\lambda = 6$), making classifiers more specialized. 
    In \cite{lev_bag}, the authors also propose using error output codes, as in \cite{output_codes}, where binary classifiers are created per class, and the final ensemble prediction is a sum of the results of classifiers per class. 
    This approach increases computational cost because the number of base learners scales with the number of classes. However, it can sometimes outperform standard Leveraging Bagging, despite the latter achieving the best average ranking in the original study \cite{lev_bag}.

    Extending the Online Boosting classifier, the authors proposed in \cite{bole} the Boosting-like Online Learning Ensemble (BOLE) algorithm. 
    Training starts with $\lambda = 1$, and classifiers are sorted by correct prediction rate (regarding correct and wrong classification weights \cite{oza_bagging}). 
    The classifier with the worst correct prediction rate is trained first. 
    If this classifier gets a correct classification, the best classifier not yet trained (in terms of correct prediction rate) receives an altered $\lambda$ for training.
    Otherwise, the worst classifier not yet trained receives it. 
    BOLE significantly increases $\lambda$ when many base learners misclassify instances, since the worst base learners influence each other until a base learner makes the right prediction. 
    The performance of the ensemble overall influences the values of $\lambda$.   

    In \cite{arf}, the authors propose the Adaptive Random Forest (ARF) algorithm, a version of the Random Forest algorithm \cite{random_forests}, where, in each split attempt, a random subset of features is considered for selecting the split with the best quality. Sampling is conducted by Poisson$(\lambda = 6)$. To deal with drifting scenarios, each classifier has two drift detectors, one for warnings and one for concept drift detection. If the warning detector triggers a change, a new classifier is trained in the background. Moreover, if the concept drift detector triggers a change, the previously trained background classifier replaces the original one. In addition to that, the classifier's votes are weighted by accuracy.

    In \cite{srp}, the authors propose the Streaming Random Patches (SRP) algorithm, a version of the Random Patches \cite{random_patches}, using the same mechanisms as in ARF, but instead of considering a random subset of features per node, all the split attempts are evaluated with a random subset of features defined at the creation of the base learner. In SRP, sampling is conducted by Poisson$(\lambda = 6)$.
    
    In~\citep{are}, the authors propose an extension to the Adaptive Random Forest (ARF) algorithm, namely Adaptive Regularized Ensemble (ARE). 
    The authors integrate an instance selection technique for the base learners of ARF, where base learners train mostly with instances that they incorrectly classify. 
    To avoid bias due to noisy instances, the authors propose a rejection strategy to train with instances that the base learner correctly classified. 
    The base classifier will train with an instance that it correctly classified after rejecting $\zeta$ (user-given) instances of the same class. 
    The authors also integrate a classifier selection technique, in which only the classifiers that perform above the average accuracy of the ensemble vote are selected.

    Another recent approach is Adaptive Random Tree Ensemble (ARTE), proposed in \citep{arte}. 
    ARTE base learners have random feature selection per node level, like ARF; however, the percentage selected is not fixed among base learners. 
    In particular, the percentage selected per base learner is drawn from a uniform distribution. 
    ARTE base learners also select the best split with random cut points, as in \cite{extra_trees}.


\section{Hoeffding Adaptive Splitting Trees Base Learners}
\label{sec:method}

As highlighted in \cite{diversity_kuncheva}, one of the key components of state-of-the-art ensembles is diversity, often induced via the distribution of a different number of copies of an instance to its base learners via random sampling with replacement, e.g., Bagging \cite{bagging_breiman}, or instance weighting, e.g., Boosting \cite{adaboost}. 
However, under adaptive splitting, ensemble members tend to generate highly correlated predictions in the early phases of the data stream, potentially undermining ensemble diversity \cite{efdt_ensemble}.

Incremental decision trees begin learning from a single root node; consequently, during the early stages of the data stream, their predictions are issued by either a majority-class classifier or a Naive Bayes model, whichever attains higher accuracy \cite{stress_testing}. 
Because every ensemble member behaves this way, the members produce nearly identical outputs, and the change detectors attached to them are therefore updated with highly similar inputs. Since instances are weighted through Poisson($\lambda$) sampling with expected value $\lambda$, the detectors exhibit similar behavior, trigger splits at closely aligned times, and ultimately yield reduced ensemble diversity. 
When the only source of differentiation among detectors is the variance introduced by Poisson weighting, split decisions are governed more by random sampling effects than by genuine performance differences or the underlying class distribution. Under these conditions, the LAST soft split criterion ($G(X_a) > 0$) becomes liable to trigger suboptimal splits.

\begin{figure}[H]
    \centering
    \includegraphics[width=0.8\linewidth]{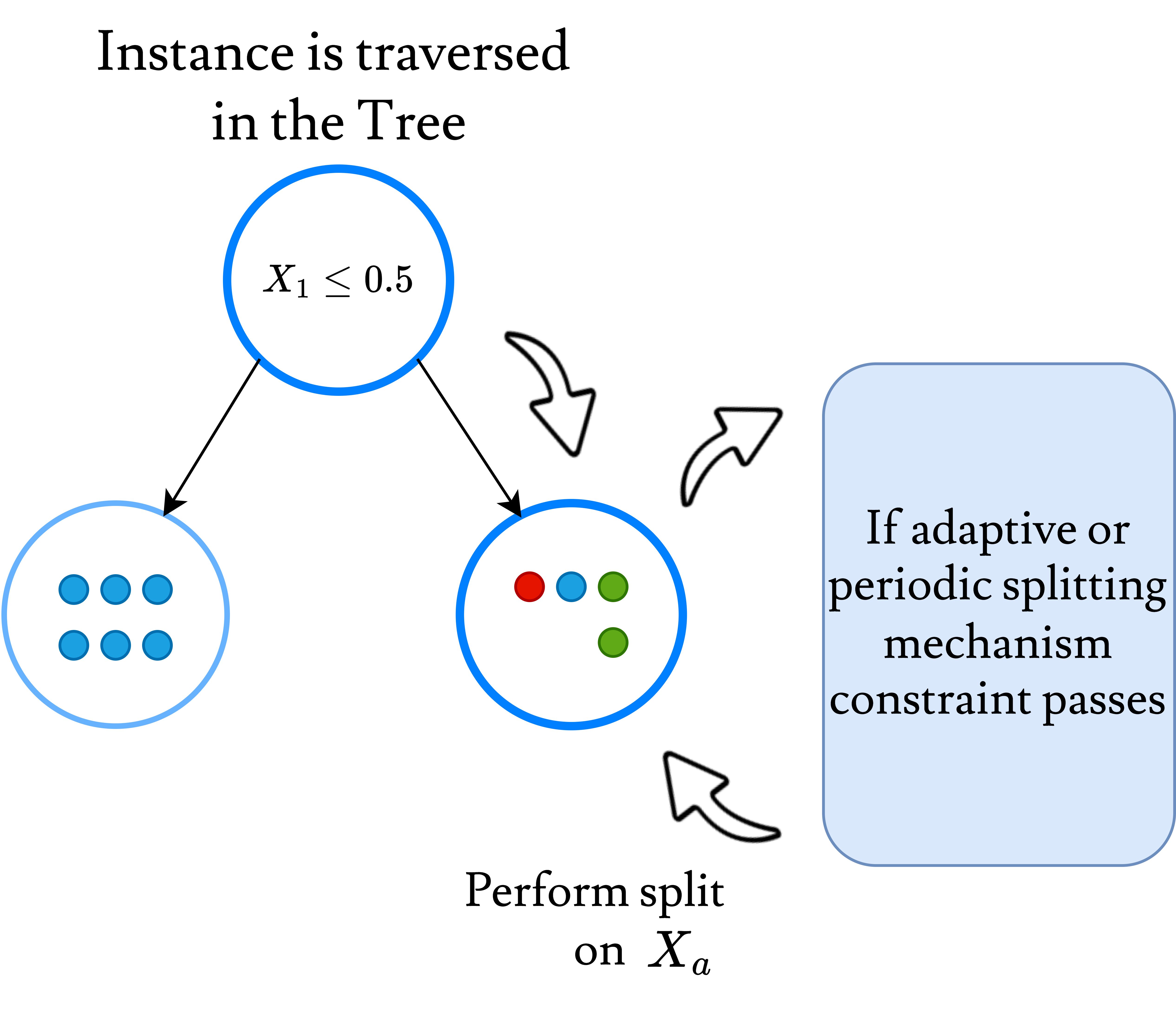}
    \caption{Training process of combining adaptive and periodic splitting mechanism}
    \label{tree:hast}
\end{figure}

To overcome this limitation, we propose two decision trees that enhance the splitting mechanism to be simultaneously adaptive, reacting to changes in the model's performance or distribution, and induce diversity among base learners of the ensemble.
Hoeffding-based Trees, such as Hoeffding Tree and EFDT, are sensitive to the number of copies they receive for training regarding splitting moment, since splitting time depends on the number of observations present in the leaf node, and present a hard constraint based on the Hoeffding bound.
Therefore, we propose a combination of the adaptive and periodic splitting mechanism, as illustrated in Figure \ref{tree:hast}.

Algorithm \ref{alg:comb_ht_efdt_last} presents the proposed methods. 
First, we propose the Hoeffding Local Adaptive Splitting Tree (HLAST). 
HLAST maintains change detectors at leaf nodes that monitor either the predictive performance or the class distribution purity, as LAST does, and periodically performs split attempts with the Hoeffding bound, as Hoeffding Trees do. 
If the change detector flags a change, and $G (X_a) > 0$ (Algorithm \ref{alg:comb_ht_efdt_last}, line 18), or $G(X_{a}) - G(X_{b}) > \sqrt{\frac{R^{2}\log(\frac{1}{\delta})}{2n}}$ (Algorithm \ref{alg:comb_ht_efdt_last}, line 14) in a periodic split attempt, the leaf is split on $X_a$.

We also propose the Extremely Fast Local Adaptive Streaming Tree (EFLAST). 
EFLAST performs splits periodically, if $G(X_a) > \sqrt{\frac{R^{2}\log(\frac{1}{\delta})}{2n}}$ (Algorithm \ref{alg:comb_ht_efdt_last}, line 16) or adaptively, if the change detector flags a change and $G(X_a) > 0$. 
We also maintain the splitting reevaluation mechanism (Algorithm \ref{alg:comb_ht_efdt_last}, line 16), where non-leaf nodes replace their branch by a new split if $G(X_{a}) - G(X_{current}) > \sqrt{\frac{R^{2}\log(\frac{1}{\delta})}{2n}}$.

Overall, the proposed trees still periodically search for the best splits and incur a computational overhead due to the use of change detectors; however, we show experimentally that such overhead is not prohibitive.

\begin{algorithm}[H]
\caption{HLAST and EFLAST: splitting strategy by combination of adaptive decision trees and HT or EFDT}\label{alg:comb_ht_efdt_last}
\begin{algorithmic}[1]
 \renewcommand{\algorithmicrequire}{\textbf{Input:}}
  \renewcommand{\algorithmicensure}{\textbf{Output:}}
\Require $S$: a data stream,\newline
$X$ : a feature, \newline
$G (\cdot)$ : a purity measure (such as Gini or Entropy), \newline
$\psi$ : a Change Detection algorithm\newline
$GP$: Grace Period (sample frequency that a leaf node will attempt to split)\newline
$DT$: Decision Tree (The constructed decision tree)\newline
splitting\_type : splitting mechanism selection between HT or EFDT (instantiating HLAST or EFLAST, respectively)
\For{$(\vec{x}_n, y_n) \in S$}
\If{splitting\_type = EFDT} 
    \State Traverse tree and ascertain that non-leaf nodes pass the reevaluation of split attempts
\EndIf  
\State Let $l \gets DT(\vec{x}_n)$ \Comment{traverse tree until leaf node}
\State $n_{l} \gets n_{l} + 1$ \Comment{increment number of samples in the leaf}
\If{  $\lnot$($l$ has samples from only one class) }
    \If{$n_{l}$ mod $GP = 0$} 
        \State Compute $G(X_{i})$ for each $X_{i} \in X_{l}$ stored in $l$ 
        \State Let $X_{a}$ be the feature with highest $G$
        \State Let $A_{b}$ be the feature with second highest $G$
        \State Let $\epsilon \gets \sqrt{\frac{R^{2}\log(\frac{1}{\delta})}{2n_l}}$ \Comment{Eq. (\ref{hoeff_ineq})}
        \If{splitting\_type = HT} 
            \State Let $\text{cond} \gets G(A_{a}) - G(A_{b}) > \epsilon$ \Comment{HLAST tree type}
        \Else 
            \State Let $\text{cond} \gets  G(A_{a}) > \epsilon$ \Comment{EFLAST tree type}  
        \EndIf         

        \If{$( (\text{cond} \lor \tau > \epsilon) \lor (l_{\psi}$ detected a change $\land  G(A_{a}) > 0)$  ) $\land ~A_{a} \neq \emptyset$ }
        \State Replace $l$ by leaf nodes that split on $X_{a}$
        \For{each leaf node $l_{i}$ from splitting on $X_{a}$}
            \State $n_{l_{i}} \gets 0$
            \State $l_{i\psi} \gets \psi$ \Comment{creates a new change detector at each leaf node}
        \EndFor
        \EndIf
        \EndIf
\EndIf   
\EndFor
\end{algorithmic}
\end{algorithm}

\section{Methodology}
\label{sec:met}

This section assesses the impact of the proposed tree models when integrated into ensembles constructed with various algorithms. The algorithms evaluated in this work were implemented in Java by extending the Massive Online Analysis (MOA) software~\citep{bifet2010moa}. 
All of our experiments are reproducible, and the source code and additional results (such as raw accuracy and F1-Score results)~are publicly available on the paper's support repository\footnote{\url{https://sites.google.com/view/last-ensemble}}.
We performed all the experiments on an Intel(R) Core(TM) i7-12700H @ 2.30 GHz with 16 GB of RAM. 
We performed experiments with 13 real-world datasets made available in \cite{souza2020challenges} and 24 synthetic datasets retrieved from \cite{kue}. Their main idea is to test how methods react to concept drift \cite{adwinbag}.
Table \ref{tab:datasets} describes the datasets used in this work. 
The datasets at the top are synthetic, while the ones at the bottom are datasets representing real-world data.

\begin{table*}[htb]
    \centering
    \caption{Description of the evaluated datasets.}
    \footnotesize
    \renewcommand{\arraystretch}{0.7}    
    \resizebox{\textwidth}{!}{
    \begin{tabular}{c|lllll}
        \hline
         Dataset & \# Samples & \# Features & \# Classes & Majority Class (\%) & Concept Drift\\ \hline
         AGR-f(7,8,9,10,9,8,7)$_{ra}$& $10^6$& 9 & 2 & 52.83 & Recurrent-Abrupt\\
         AGR-f(7,8,9,10,9,8,7)$_{rg}$& $10^6$& 9 & 2 & 52.83 & Recurrent-Gradual\\
         AGR-f(1,2,3,4,5,6,7,8,9,10)$_{a}$& $10^6$& 9 & 2 & 52.83 & Abrupt\\
         AGR-f(10,9,8,7,6,5,4,3,2,1)$_{a}$& $10^6$& 9 & 2 & 52.83 & Abrupt\\
         AGR-f(1,2,3,4,5,6,7,8,9,10)$_{g}$& $10^6$& 9 & 2 & 52.83 & Gradual\\
         AGR-f(10,9,8,7,6,5,4,3,2,1)$_{g}$& $10^6$& 9 & 2 & 52.83 & Gradual\\
         SEA-f(1,2,3,4,3,2,1)$_{ra}$& $10^6$& 3 & 2 &59.91 & Recurrent-Abrupt\\
         SEA-f(1,2,3,4,3,2,1)$_{rg}$& $10^6$& 3 & 2 &59.91 & Recurrent-Gradual\\
         SEA-f(1,2,3,4)$_{a}$& $10^6$& 3 & 2 &59.91 & Abrupt\\
         SEA-f(1,2,3,4)$_{g}$& $10^6$& 3 & 2 &59.91 & Gradual\\
         SEA-f(4,3,2,1)$_{a}$& $10^6$& 3 & 2 &59.91 & Abrupt\\
         SEA-f(4,3,2,1)$_{g}$& $10^6$& 3 & 2 &59.91 & Gradual\\
         LED-f(7,5,3,1,3,5,7)$_{ra}$&$10^6$& 24 & 10 & 10.28 & Recurrent-Abrupt\\
         LED-f(7,5,3,1,3,5,7)$_{rg}$&$10^6$& 24 & 10 & 10.28 & Recurrent-Gradual\\
         LED-f(1,3,5,7)$_{a}$&$10^6$& 24 & 10 & 10.28 & Abrupt\\
         LED-f(1,3,5,7)$_{g}$&$10^6$& 24 & 10 & 10.28 & Gradual\\
         LED-f(7,5,3,1)$_{a}$&$10^6$& 24 & 10 & 10.28 & Abrupt\\
         LED-f(7,5,3,1)$_{g}$&$10^6$& 24 & 10 & 10.28 & Gradual\\
         RBF$_{s}$& $10^6$&10 & 5 & 30.01 & Incremental\\
         RBF$_{m}$& $10^6$&10 & 5 & 30.01 & Incremental\\
         RBF$_{f}$& $10^6$&10 & 5 & 30.01 & Incremental\\
         HYPER$_{s}$& $10^6$&10 & 2 & 50 & Incremental \\
         HYPER$_{m}$& $10^6$&10 & 2 & 50 & Incremental \\
         HYPER$_{f}$& $10^6$&10 & 2 & 50 & Incremental \\

         \hline
         Outdoor & 4,000 & 21 & 40 & 4.11 & Unknown \\
         Elec & 45,312 & 8 & 2 & 57.41 & Unknown \\
         Rialto & 82,250 & 27 & 10 & 10 & Unknown \\ 
         Airlines & 539,383& 7 & 2 & 55.47 & Unknown \\
         CoverType & 581,012 & 54 & 7 & 48.75 & Unknown \\
         Nomao & 34,465 & 119 & 2 & 71.44 & Unknown \\
         Poker & 829,201 & 10 & 10 & 47.78 & Unknown \\        
         NOAA & 18,158 & 8 & 2 & 69.74 & Unknown \\   
         INSECTS$_{a}$ & 52,848 & 33 & 6 & 16.07 & Abrupt\\ 
         INSECTS$_{i}$ & 57,018 & 33 & 6 & 11.56 & Incremental \\
         INSECTS$_{g}$ &  24,150 & 33 & 6 & 15.76 & Gradual \\ 
         Asfault & 8,066 & 62 & 5 & 55.59  & Unknown \\
         LADPU & 22,950 & 96 & 10 & 10 & Unknown\\
         \hline         
    \end{tabular}
    }
    \label{tab:datasets}
\end{table*}

     The synthetic dataset generators are described as follows.

     \paragraph{AGRAWAL \cite{agrawal}} The AGRAWAL generator simulates a loan-approval scenario with nine features that represent applicant attributes: \textit{salary} (numerical, 0--150K), \textit{commission} (numerical, 0--75K), \textit{age} (numerical, 20--80), \textit{eloan} (boolean), \textit{car} (nominal, 1--20), \textit{zipcode} (nominal, 0--8), \textit{hvalue} (numerical, 50K--600K), \textit{hyears} (numerical, 1--30), and \textit{loan} (numerical, 0--500K). Ten classification functions encode distinct loan-approval rules over these attributes, mapping feature combinations to a binary label (approved or denied). These functions are progressively more complex from $f_1$ to $f_{10}$: the earlier ones depend on a single attribute with simple cut points, while the later ones combine several attributes through compound conditions and derived quantities. For instance, $f_1$ assigns the label using only \textit{age}, approving applicants younger than 40 or at least 60 years old and denying the remaining middle-age range. In contrast, $f_{10}$ is the most elaborate rule, computing a derived \textit{disposable income} from \textit{salary}, \textit{commission}, and \textit{loan} together with an equity term that depends on \textit{hvalue} and \textit{hyears}, and approving the loan only when this disposable income is positive. Concept drift is simulated by switching between these functions. Following the table, we use a recurrent variant cycling through $\{f_7,f_8,f_9,f_{10},f_9,f_8,f_7\}$ (6 drifts) and monotone ascending $\{f_1,\dots,f_{10}\}$ and descending $\{f_{10},\dots,f_1\}$ variants (9 drifts each). 

     \paragraph{SEA \cite{street2001sea}} The SEA generator produces three numerical features $f_{1}, f_{2}, f_{3}$ uniformly sampled from $[0,10)$, where $f_{3}$ is entirely irrelevant to the class label, making this a three-feature problem with an embedded irrelevant dimension. The class label is binary and is determined by a threshold applied to the sum of the two relevant features. Four concepts are defined as $c_1\colon (f_{1} + f_{2} \leq 8)$, $c_2\colon (f_{1} + f_{2} \leq 9)$, $c_3\colon (f_{1} + f_{2} \leq 7)$, and $c_4\colon (f_{1} + f_{2} \leq 9.5)$: an instance receives class 1 if the condition for the active concept holds, and class 0 otherwise. The recurrent variant cycles through concepts in the order $\{c_1, c_2, c_3, c_4, c_3, c_2, c_1\}$ (6 drifts), while the monotone abrupt and gradual variants follow $\{c_1,c_2,c_3,c_4\}$ or $\{c_4,c_3,c_2,c_1\}$ (3 drifts each). Gradual transitions are controlled by a window in which instances are drawn from both the outgoing and incoming concept with a probability that changes linearly.

     \paragraph{LED \cite{cart}} The LED generator simulates reading a seven-segment LED display showing digits 0--9, where each digit is encoded by 7 boolean features corresponding to the active segments. An additional 17 boolean features are appended and are entirely irrelevant to the classification task, making feature selection non-trivial. Furthermore, each boolean feature is independently flipped with a 10\% probability, introducing label noise. Concept drift is simulated by progressively masking or unmasking the originally relevant features: as the number of relevant features decreases, fewer segments are correctly observed and the classification boundary becomes harder to learn. In our experimental setup the order of relevant feature counts across concepts is $\{7, 5, 3, 1, 3, 5, 7\}$ for the recurrent variant (LED$_{ra}$ and LED$_{rg}$, 6 drifts) and $\{1,3,5,7\}$ or $\{7,5,3,1\}$ for the monotone variants (LED$_a$ and LED$_g$, 3 drifts each). Abrupt variants transition instantaneously between concepts, whereas gradual variants mix instances from adjacent concepts over a transition window.

    \paragraph{RBF \cite{bifet2010moa}} The Radial Basis Function (RBF) generator creates a Gaussian mixture model over 10 features and 5 classes, in which each of the 50 randomly initialised centroids is associated with a standard deviation, a weight, and a class label; instances are drawn by selecting a centroid with probability proportional to its weight and sampling from the corresponding Gaussian. Incremental concept drift is introduced by continuously moving the centroids through the feature space at a fixed speed, so that the decision boundaries shift gradually over time rather than switching abruptly. Three speed variants are evaluated: RBF$_s$ (slow, speed $= 10^{-5}$), RBF$_m$ (moderate, speed $= 10^{-4}$), and RBF$_f$ (fast, speed $= 10^{-3}$). The increasing speed makes it progressively harder for learners to track the drifting class regions.

    \paragraph{HYPER \cite{hulten2001cvfdt}} A hyperplane in $\mathbb{R}^d$ is a $(d-1)$-dimensional affine subspace defined by $\sum_{i=1}^{d} w_i x_i = w_0$, which partitions the space into two half-spaces and thus provides a natural binary decision boundary. Instances are generated uniformly at random in $[0,1]^d$, and their class is determined by which side of the hyperplane they fall on. Incremental concept drift is simulated by continuously rotating the hyperplane: at each step, a random weight $w_i$ is perturbed by an amount proportional to the magnitude parameter $\sigma$, and all weights are subsequently re-normalised. Three speed variants are evaluated: HYPER$_s$ (slow, $\sigma = 10^{-4}$), HYPER$_m$ (moderate, $\sigma = 10^{-3}$), and HYPER$_f$ (fast, $\sigma = 10^{-2}$). All variants were configured with $d = 10$ features.


In contrast to the experiments conducted in \cite{efdt_ensemble}, all ensembles were set with 100 base learners, as used in recent state-of-the-art ensembles \cite{arf,srp}. 
All parameters from ensembles were set to default as implemented in the MOA framework. In addition to the ensembles evaluated in \cite{efdt_ensemble}, we also experimented with SRP~\citep{srp}, ARE~\citep{are}, and ARTE~\citep{arte}. 
We acknowledge SGBT ~\citep{sgbt}, a method that adapts gradient boosting trees for streaming scenarios; however, many datasets ran for a week constantly demanding memory swap and presented a high computational cost, and were therefore excluded from the experiments. We also chose not to include in our experiments ensemble algorithms specifically designed for imbalanced contexts, such as ROSE \cite{rose}, KUE \cite{kue}, and others cited in \cite{survey_imb_cano}, opting instead to focus on general-purpose ensemble methods.
Only the best state-of-the-art methods are evaluated for visualization convenience, including Leveraging Bagging, ARF, SRP, ARE, and ARTE, based on the results reported in \cite{arf, srp, arte}.

As in \citep{efdt}, we assess the predictive performance obtained with a test-then-train validation strategy, where every instance is used first for testing and then for training, known as Prequential evaluation \cite{eval_streams_gama}. 
All metrics are averaged over 20 evenly spaced points of the stream, as more points did not change results drastically.

To statistically compare the results obtained by the ensembles of base learners, we performed a Friedman test with a significance level of 1\% (p-value $< 0.01$) and a pairwise one-sided Wilcoxon signed-rank post-hoc test with a Holm correction ~\citep{garcia_herrera, benavoli}. 
We visually represent the results of this test using a diagram as in ~\citep{demvsar2006statistical}. 
In this diagram, the methods are sorted according to their average ranking output by the statistical test. 
The methods connected by a line do not present statistically significant differences.

Time processing and memory efficiency are key components of data stream mining algorithms, as cited earlier and in \cite{bifet2010moa}. 
Although not covered in the experimentation presented in \cite{efdt_ensemble}, in this work we evaluate CPU-Time, memory usage (peak RAM, in MB, throughout time), and the mean tree size (number of nodes) in the ensemble.
For these metrics, we provide violin plots sorted by median.

LAST variants that incorporate data distribution monitoring at the leaf nodes are indicated by the suffix ``$D$'', as in LAST$_{D}$, HLAST$_{D}$ and EFLAST$_{D}$.
The detector used in all LAST versions was HDDM$_{A}$~\cite{hddm}, given the analysis done in \cite{kais_daniel}.
It is important to clarify the relationship between this work and \cite{kais_daniel}, which evaluates LAST exclusively as a monolithic decision tree and reports no ensemble experiments. 
Two of its contributions are directly relevant here. 
First, \citeauthor{kais_daniel} \citep{kais_daniel} conjecture that LAST$_{D}$, which monitors class-distribution purity rather than error rate, may suit ensemble use better than LAST: because of the instance sampling process, leaf nodes are expected to undergo more frequent distributional changes, which could lead LAST$_{D}$ to induce greater diversity among base learners. 
We investigate this hypothesis in the present work. 
Second, among the change detectors evaluated, HDDM$_{A}$, MDDM$_{A}$, and ADWIN stand out: they produce trees as accurate as the DDM-type detectors (DDM, EDDM, RDDM) while remaining more computationally efficient, owing to stricter detection bounds and fewer false positives. 
We adopt HDDM$_{A}$, which achieved the best overall ranking, though MDDM$_{A}$ and ADWIN performed comparably.

\section{Results}
\label{sec:res} 

This section presents results for experimentation conducted to assess the proposed methods.
First, Section \ref{subsec:benchmark} presents a benchmark for ensembles and base learners. 
Next, Section \ref{subsec:tree_size} presents a comparison of tree sizes, while Section \ref{subsec:comp_cost} discusses the computational cost of the methods. 
Finally, Section \ref{subsec:concept_drift} analyzes the F1-Score of methods in concept drifting scenarios.


\subsection{Benchmark}
\label{subsec:benchmark}

This section benchmarks the proposed trees against Hoeffding Trees (HT), EFDT, and the original adaptive trees LAST and LAST$_D$ as base learners of the five ensembles evaluated in this work, where we report F1-Score. 
We discuss the behaviour of each base tree per dataset, so that the properties of the data that favour the proposed trees become explicit. 
The raw F1-Score and accuracy values for every combination of ensemble and base learner are available in the paper's repository.

\subsubsection{Best-performing base tree}
\label{best_tree}

\begin{figure}[H]
    \centering
    \includegraphics[width=\linewidth]{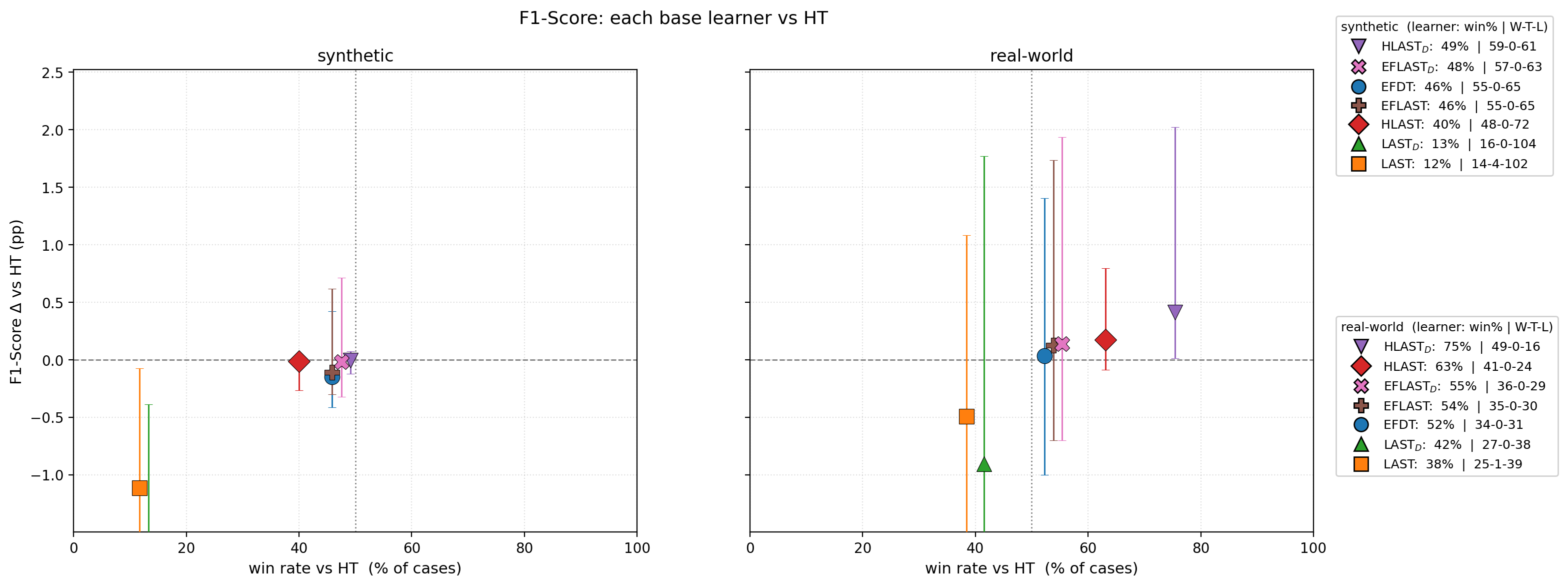}
    \caption{F1-Score of each base learner against HT, reported as the percentage of cases in which the base learner wins over HT (win rate) and the F1-Score difference quantiles in percentage points, for synthetic (left) and real-world (right) data}
    \label{fig:effect_tree}
\end{figure}

Figure \ref{fig:effect_tree} summarizes how each base tree compares against a standard Hoeffding Tree across all ensembles. 
The horizontal axis reports the percentage of cases in which the base learner wins over HT, while the vertical axis reports the F1-Score difference to HT in percentage points, where the marker is the median difference across all datasets and ensembles and the vertical bars span the lower and upper quantiles of that difference. 
In real-world data, HLAST$_D$ and HLAST are among the best-performing trees, winning over HT in $75\%$ and $63\%$ of the cases, respectively, ahead of EFDT and EFLAST, and largely ahead of the original adaptive trees LAST and LAST$_D$, which win in only $38\%$ and $42\%$ of the cases. 
Considering the quantiles, HLAST$_D$ and HLAST keep a median difference at or above zero, and their upper quantile reaches above two and one percentage points, respectively, meaning that on the favorable datasets the gains are substantial, while their lower quantile stays close to zero, meaning they rarely lose much when they do not win. 
The original adaptive trees show the opposite behaviour, with a median difference below zero, which confirms that LAST and LAST$_D$ are weak ensemble base learners, while combining the adaptive mechanism with the periodic Hoeffding split recovers and surpasses the performance of HT.

\begin{figure}[H]
    \centering
    \includegraphics[width=\linewidth]{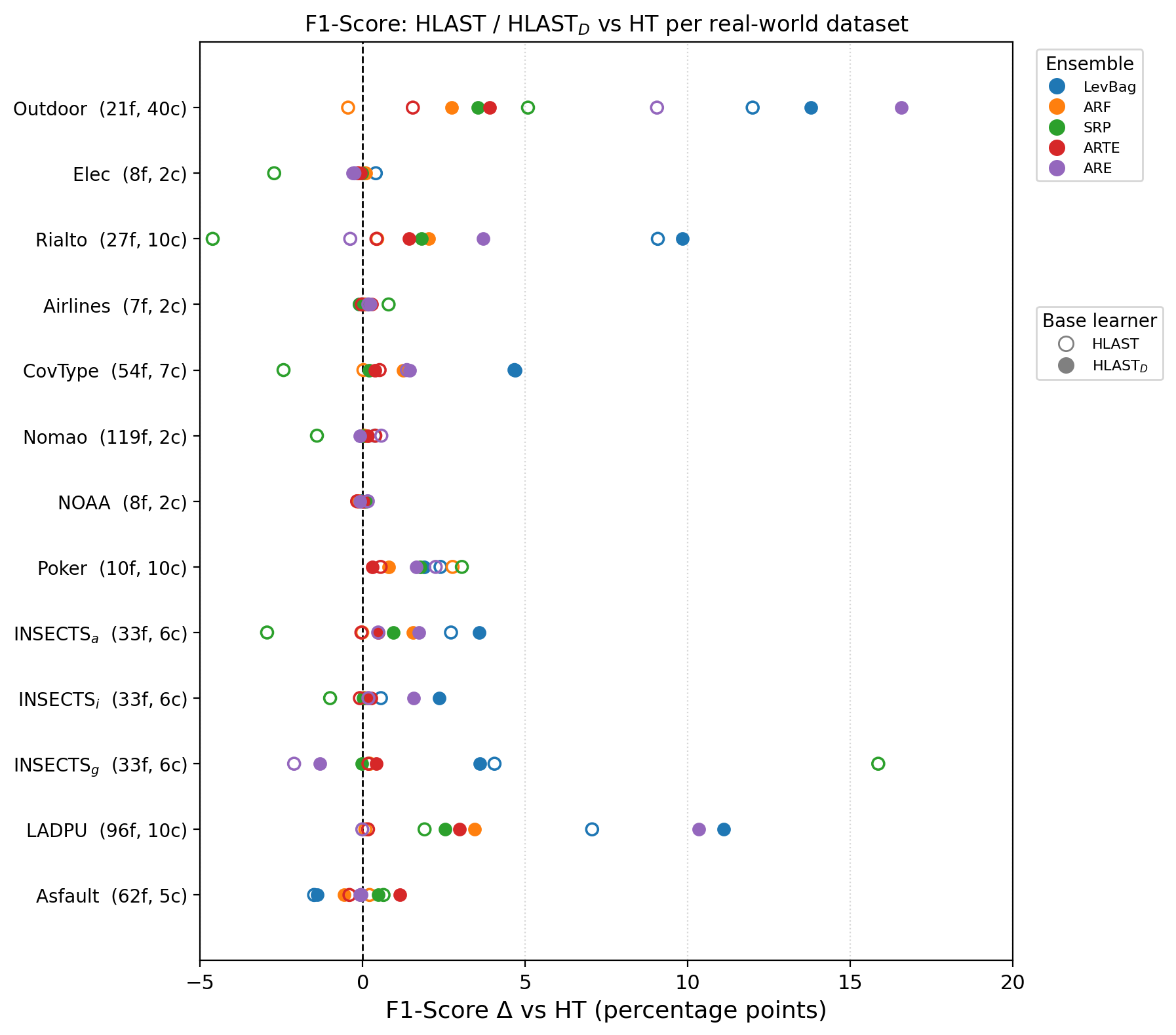}
    \caption{F1-Score difference between HLAST/HLAST$_D$ and HT in real-world data for all ensembles evaluated}
    \label{fig:dotstrip_real}
\end{figure}

Figure \ref{fig:dotstrip_real} details this difference per dataset, with each point colored by the ensemble that produced it. The gains of the proposed trees concentrate on datasets with a larger number of classes, such as Outdoor (40 classes), Rialto, Poker and LADPU (10 classes), CoverType (7 classes) and the INSECTS variants (6 classes), reaching up to $16$ percentage points in LADPU and around $10$ points in Outdoor and Rialto. 
The highest gains come from Leveraging Bagging and ARE, the ensembles that grow the largest trees (Table \ref{mean_tree_size}), so the adaptive splitting strategy has the most pronounced effect on them, but the remaining ensembles also present positive gains in general, and the proposed trees produce higher predictive quality across the board. 
On binary datasets such as Electricity, Airlines, NOAA and Nomao, the difference stays close to zero. This behaviour follows directly from the description of the proposed trees. 
With more classes, a leaf node takes longer to become pure and offers more opportunities to split, so letting the change detector decide when to split, on top of the periodic Hoeffding attempts, produces larger and more diverse trees (Table \ref{mean_tree_size}). 
On binary problems the concept is learned quickly and the additional adaptive splitting remains mostly idle, leaving little room for improvement.

The datasets on which the proposed trees perform best are also the most challenging, according to the streaming benchmark in \cite{souza2020challenges}.
That work characterizes real-world streams by their temporal dependence and shows that, due to autocorrelation, a naive No-Change baseline---one that simply predicts the label of the most recent instance---can already achieve strong performance, rendering several commonly used benchmark datasets deceptively easy.
This is not the case for Rialto, LADPU, and Asfault, where the proposed trees achieve their highest gains: these datasets lack such autocorrelation and thus require the model to genuinely learn the underlying concept.
Even on datasets with autocorrelation, such as Electricity and CoverType, the ensembles outperform the No-Change baseline by a wide margin. In terms of F1-Score, ARTE and HLAST$_{D}$ reach $91.33\%$ against $84.43\%$ on Electricity and $86.60\%$ against $76.41\%$ on CoverType, confirming that the reported gains reflect genuine concept learning rather than artifacts of easy data.

Figure \ref{fig:dotstrip_real} further shows that the two proposed trees do not behave identically across ensembles.
Under SRP, HLAST is the weakest of the three base learners, with a standard HT outperforming it in roughly $70\%$ of the cases.
The reason lies in how SRP operates: it assigns each base learner a fixed random subset of features at creation, so some learners are inevitably built on weak or uninformative subsets.
Because HLAST relies on a soft adaptive split driven by the error rate, these learners continue splitting even when their features are poor, and since SRP weights votes by accuracy, the resulting large but weakly featured trees bias the weighted voting and degrade the ensemble.
HLAST$_{D}$ avoids this pitfall: by monitoring class-distribution purity rather than the error rate, its growth is not driven by fluctuations of the error detector.
HLAST$_{D}$ is therefore the natural choice to pair with SRP, whereas HLAST is better suited to ensembles such as ARTE.

\begin{figure}[H]
    \centering
    \includegraphics[width=\linewidth]{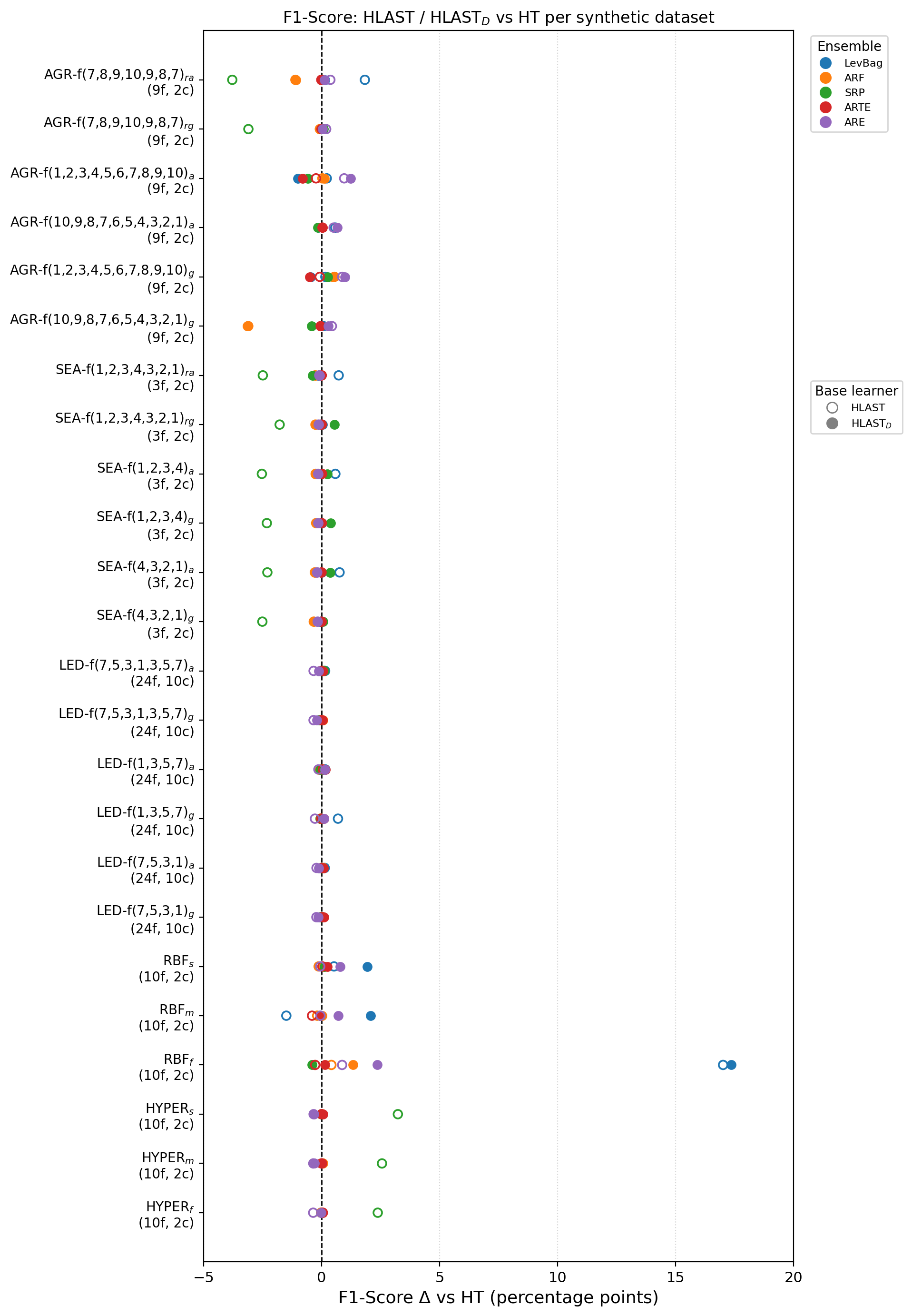}
    \caption{F1-Score difference between HLAST/HLAST$_D$ and HT in synthetic data for all ensembles evaluated}
    \label{fig:dotstrip_synth}
\end{figure}

Figure \ref{fig:dotstrip_synth} shows that on synthetic data the proposed trees have little effect, with the F1-Score difference to HT staying close to zero for all ensembles. The synthetic generators used in the benchmark configure simple linear concepts, as in AGRAWAL, SEA and HYPER, or binary patterns, as in LED, and their main purpose is to test how methods react to concept drift. As further discussed in Section \ref{subsec:concept_drift}, the ensembles react to drift in a very similar way, and the simple concepts are learned fast by the trees, so growing them further with adaptive splitting does not change the results in a meaningful way. The impact of the proposed trees is therefore more evident on real-world data, where the concepts are more complex.

\subsubsection{Best-performing ensemble}
\label{best_ensemble}

\begin{figure}[H]
    \centering
    \includegraphics[width=\linewidth]{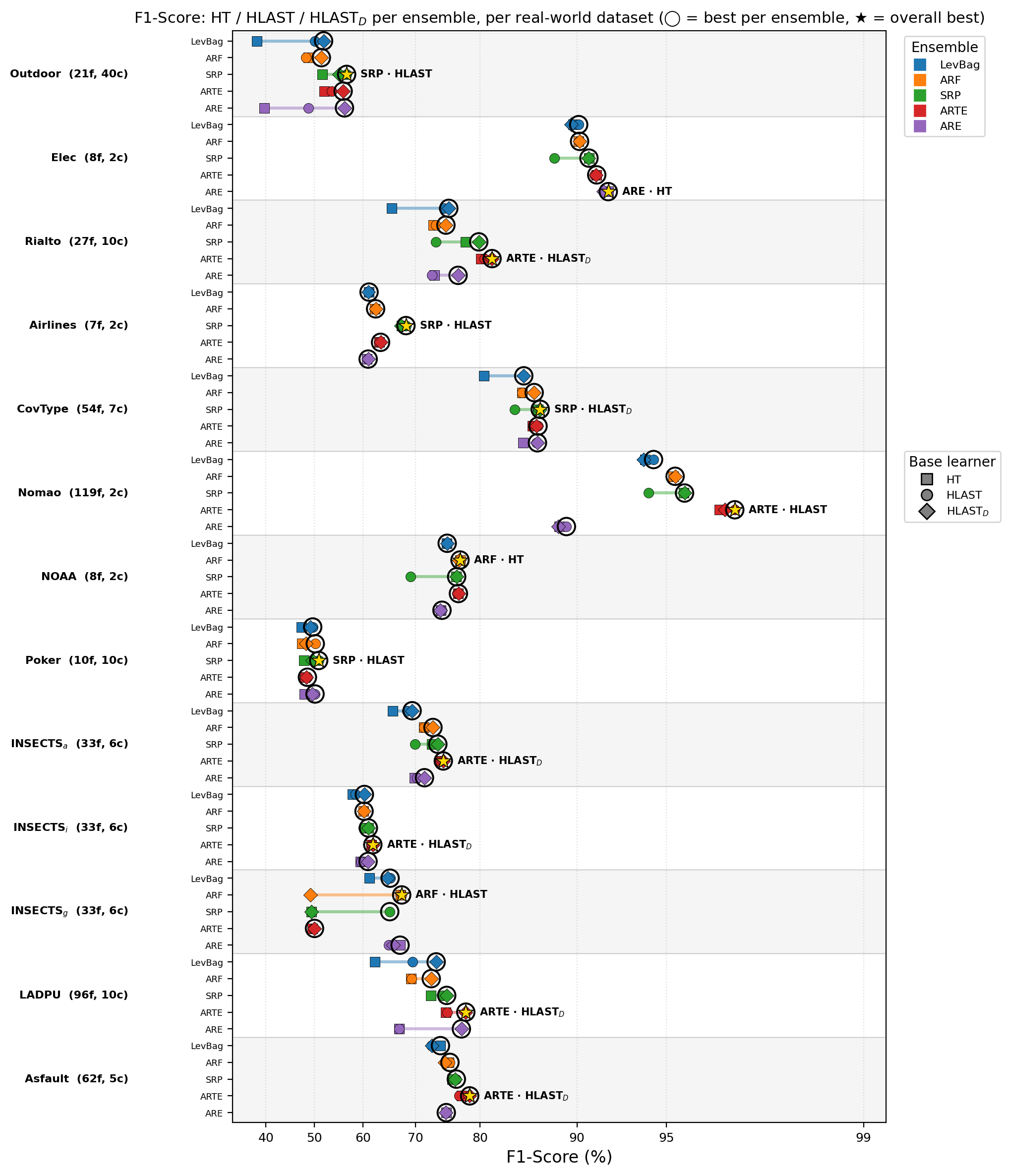}
    \caption{Raw F1-Score for HT/HLAST/HLAST$_D$ in real-world data for all ensembles evaluated}
    \label{fig:raw_f1_real}
\end{figure}

Figure \ref{fig:raw_f1_real} reports the raw F1-Score of HT, HLAST, and HLAST$_{D}$ for every ensemble, marking the best base learner within each ensemble and the overall best result per dataset.
The proposed trees provide the overall best result on most real-world datasets, and almost all of these wins occur on the multi-class datasets, in line with the per-tree analysis.
Specifically, ARTE with HLAST$_{D}$ delivers the best result on Rialto, the INSECTS variants, LADPU, and Asfault; ARTE with HLAST leads on Nomao; and SRP with HLAST or HLAST$_{D}$ leads on Outdoor, Airlines, CoverType, and Poker.

Considering the results as a whole, the proposed trees achieve the strongest performance, and we recommend ARTE as the ensemble to pair with them.
ARTE combines a per-learner random feature percentage with random cut points, which already induces high diversity; the proposed trees build on this by splitting at more informed moments rather than blindly at fixed intervals.
The combination is particularly effective on real-world data, where ARTE with HLAST$_{D}$ outperforms ARTE with HT on almost every dataset and yields the strongest and most consistent results across the entire benchmark.

Figure \ref{fig:raw_f1_synth} shows that, on synthetic data, the differences between base learners are small and the best result alternates among ensembles and trees, with Leveraging Bagging leading on LED, SRP on HYPER, and ARTE on SEA.
This reinforces the point that synthetic concepts do not clearly separate the methods, and that the advantage of the proposed trees (and of ARTE in particular) stems mainly from the more complex real-world data.

\begin{figure}[H]
    \centering
    \includegraphics[width=\linewidth]{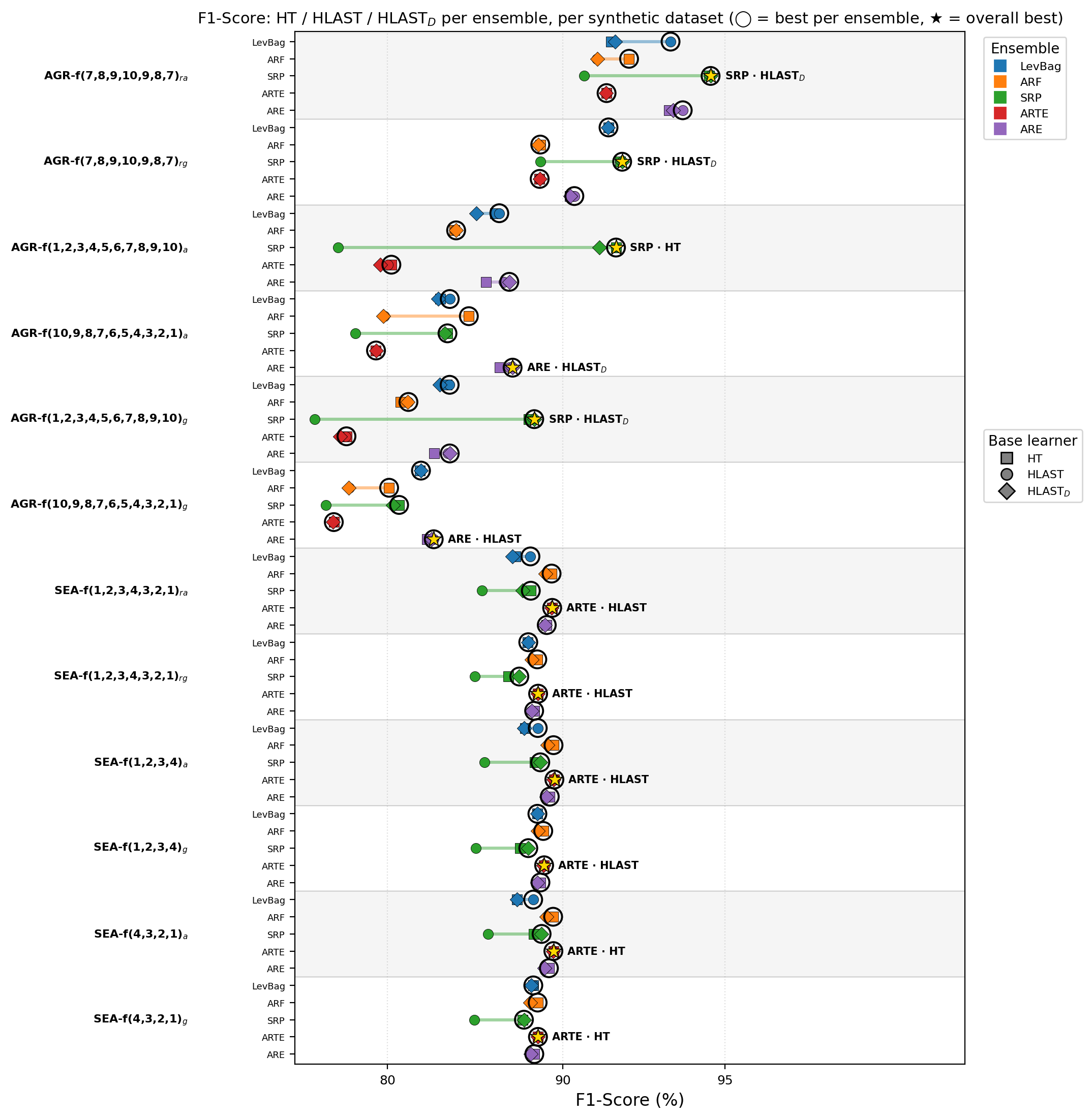}
    \caption{Raw F1-Score for HT/HLAST/HLAST$_D$ in synthetic data for all ensembles evaluated}
    \label{fig:raw_f1_synth}
\end{figure}

\begin{figure}[H]
    \ContinuedFloat
    \centering
    \includegraphics[width=\linewidth]{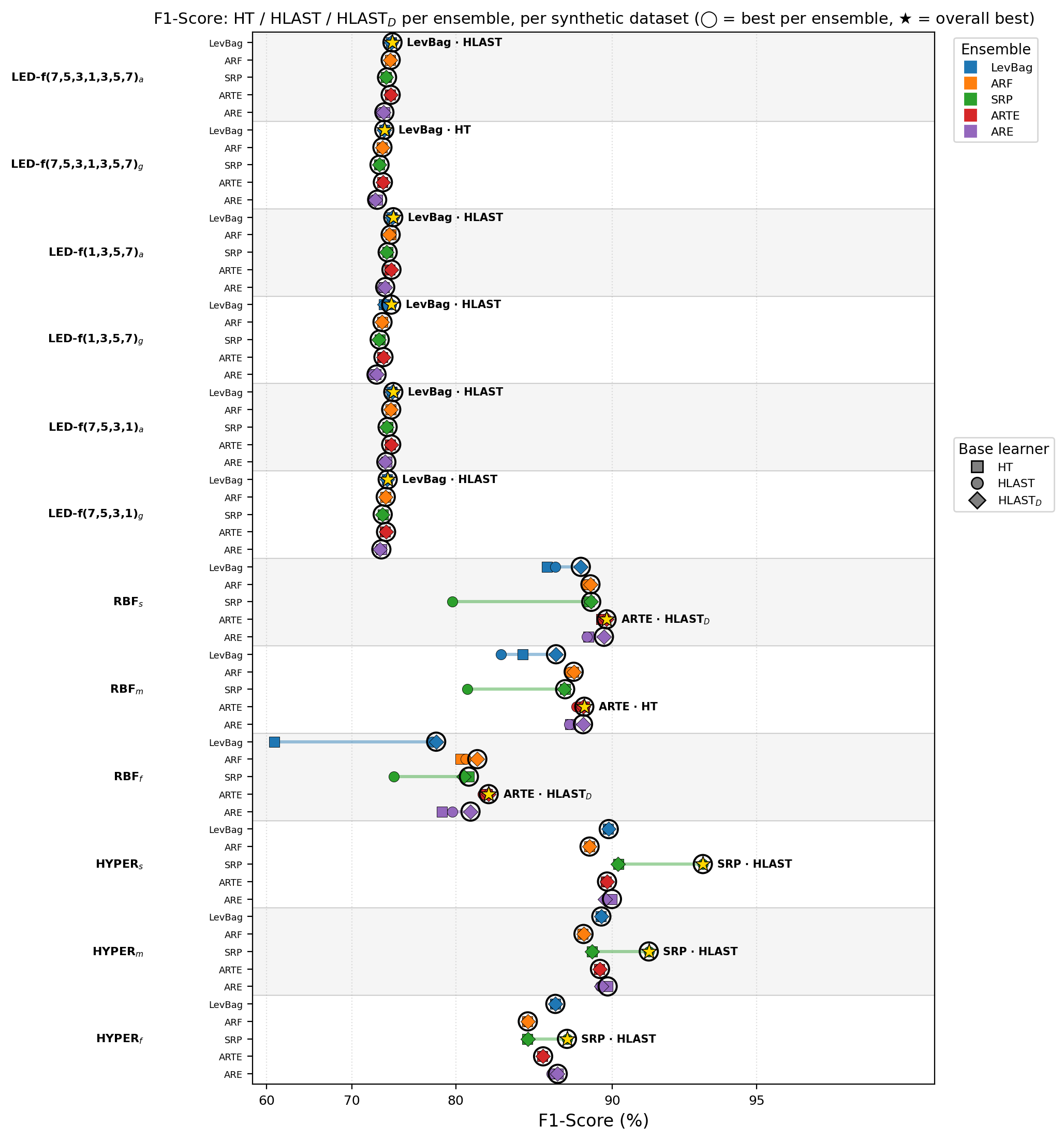}
    \caption{(Continued) Raw F1-Score for HT/HLAST/HLAST$_D$ in synthetic data for all ensembles evaluated}
    \label{fig:raw_f1_synth_part2}
\end{figure}

The main points observed in this section were:
\begin{itemize}
     \item [1.] HLAST and HLAST$_D$ are among the best-performing base trees on real-world data, surpassing HT and EFDT, while the original adaptive trees LAST and LAST$_D$ are the weakest as ensemble base learners.
     \item [2.] The proposed trees yield larger gains on datasets with more classes, where leaves stay impure longer and the adaptive splitting has more room to act, and have little effect on binary and synthetic data, where the simpler concepts are learned fast.
     \item [3.] HLAST$_D$ is the most suitable tree for SRP, as HLAST lets learners built on weak feature subsets keep growing and biases SRP's accuracy-weighted voting.
     \item [4.] The proposed trees achieve the best performance overall, and ARTE with HLAST$_D$ is the recommended combination, providing the strongest and most consistent results on real-world data.
\end{itemize}

\subsection{Tree Size}
\label{subsec:tree_size}

Table \ref{mean_tree_size} shows the mean tree size and standard deviation of ARF decision trees, which had similar behavior in other ensembles as well. As expected, LAST presented lower standard deviation compared to other base learners, showing that LAST presents low diversity as a base learner of the ensemble. EFDT presented lower tree size compared to Hoeffding-based Trees due to the reevaluation process. However, lower tree size does not imply lower memory cost, as EFDT needs to store data at non-leaf nodes.

    \begin{table}[!htb]
    \centering
    \scriptsize			
    \caption {Mean and standard deviation of tree size of ARF base learners}
    \label{mean_tree_size}
    \begin{tabular}{llll} 
    \hline 
    \textbf{HT}&\textbf{EFDT}&\textbf{LAST}&\textbf{LAST}$_{\mathbf{D}}$\\ 
\hline 
683,91 $\pm$ 3836.14 & 344.20 $\pm$ 717.44 & 40.70 $\pm$ 22 & 10.27 $\pm$ 9.22\\
\hline  
\textbf{HLAST}&\textbf{HLAST}$_{\mathbf{D}}$&\textbf{EFLAST}&\textbf{EFLAST}$_{\mathbf{D}}$\\
\hline 
1062.42 $\pm$ 8062.94 & 1106.77 $\pm$ 8109.61 & 347.15 $\pm$ 812.37 & 359.33 $\pm$ 782.68\\ 
\hline 
    \end{tabular} 
    \label{arf:desv_pad}
    \end{table}

\begin{figure}[h]
\centering
\begin{subfigure}[b]{1\textwidth}
          \includegraphics[width=\textwidth]{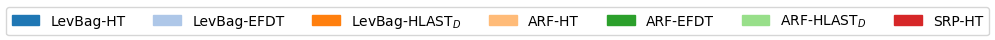}
\end{subfigure} \\
\begin{subfigure}[b]{1\textwidth}
          \includegraphics[width=\textwidth]{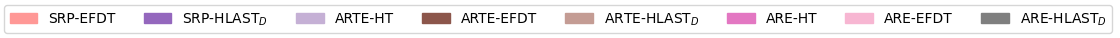}
\end{subfigure} \\
\centering
\includegraphics[width=0.8\textwidth]{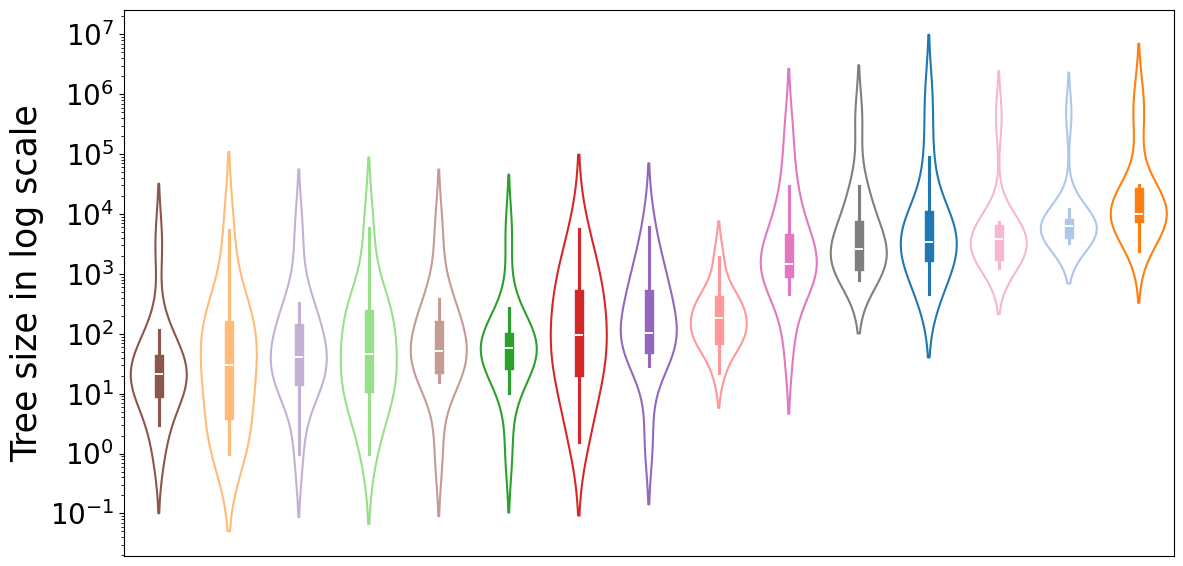}
\includegraphics[width=0.8\textwidth]{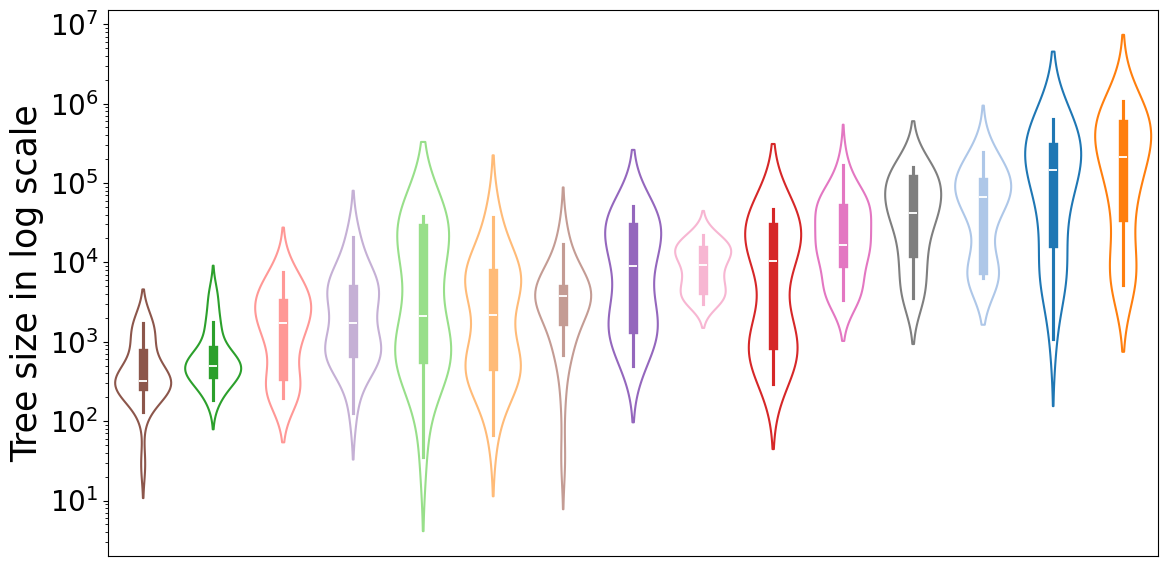}
\caption{Distribution of average tree size of the ensembles with HT, EFDT and best-ranking reported proposed method per ensemble as base learners for real-world (top) and synthetic data (down)}
\label{tree_size}
\end{figure}

Fig. \ref{tree_size} shows the distribution of average tree size in real-world and synthetic datasets. In real-world datasets, ARTE with HLAST$_{D}$ presented higher median and third quartile compared to ARTE with HT, but still presented lower third quartile, median and first quartile compared to SRP with HT. 

ARE and LevBag presented the largest tree sizes among all ensembles and their respective base learners. The high tree size of ARE could be explained by the input of the change detectors, which depend mostly on instances that were misclassified and used for training by the base learner. An update of the change detector with all the instances, independently of rejection, could mitigate this. In synthetic data, ARTE with HLAST$_{D}$ also presented a higher median and third quartile than ARTE with HT and lower third quartile, median and first quartile compared to SRP with HT.

The main points observed in this section were:
\begin{itemize}
     \item [1.] LAST presents a low standard deviation in tree size, showing lower diversity of the base learner as a member of the ensemble.
     \item [2.] ARE base learners' change detectors receive mostly instances incorrectly classified by the base learner, and updating the change detector with all instances could help the change detector to track the performance of the base learner, independently of training or not with the instance.   
     \item  [3.] ARTE with HLAST$_{D}$ presents a higher tree size compared to ARTE with HT, but still a lower tree size compared to SRP with HT.
\end{itemize}

\newpage

\subsection{Computational Cost}
\label{subsec:comp_cost}

\begin{figure}[h]
\centering
\begin{subfigure}[b]{1\textwidth}
          \includegraphics[width=\textwidth]{legend_1.png}
\end{subfigure} \\
\begin{subfigure}[b]{1\textwidth}
          \includegraphics[width=\textwidth]{legend_2.png}
\end{subfigure} \\
\centering
\includegraphics[width=0.8\textwidth]{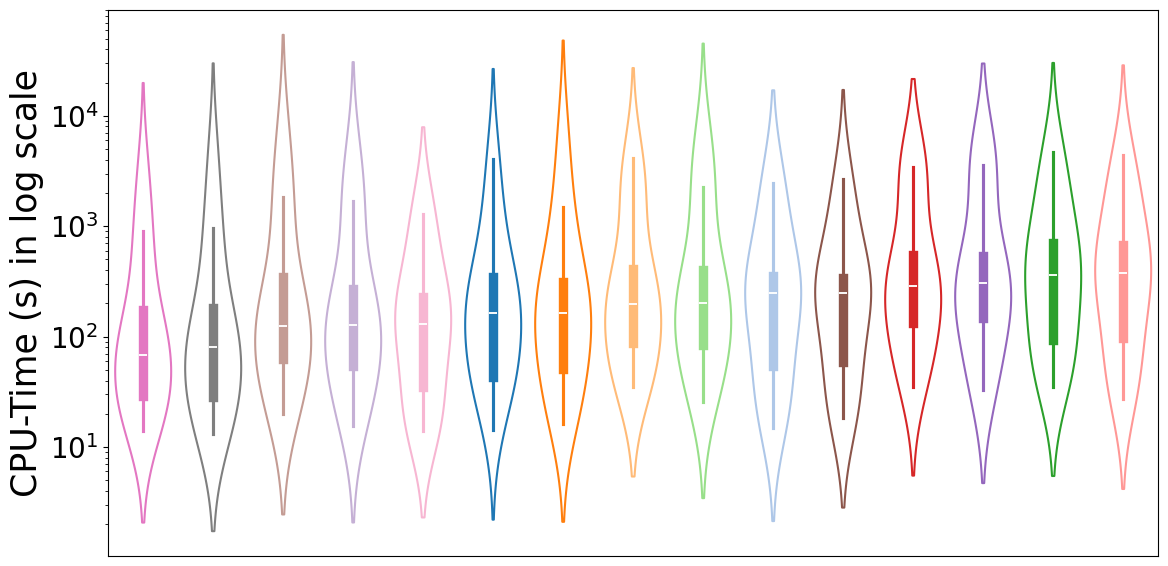}
\includegraphics[width=0.8\textwidth]{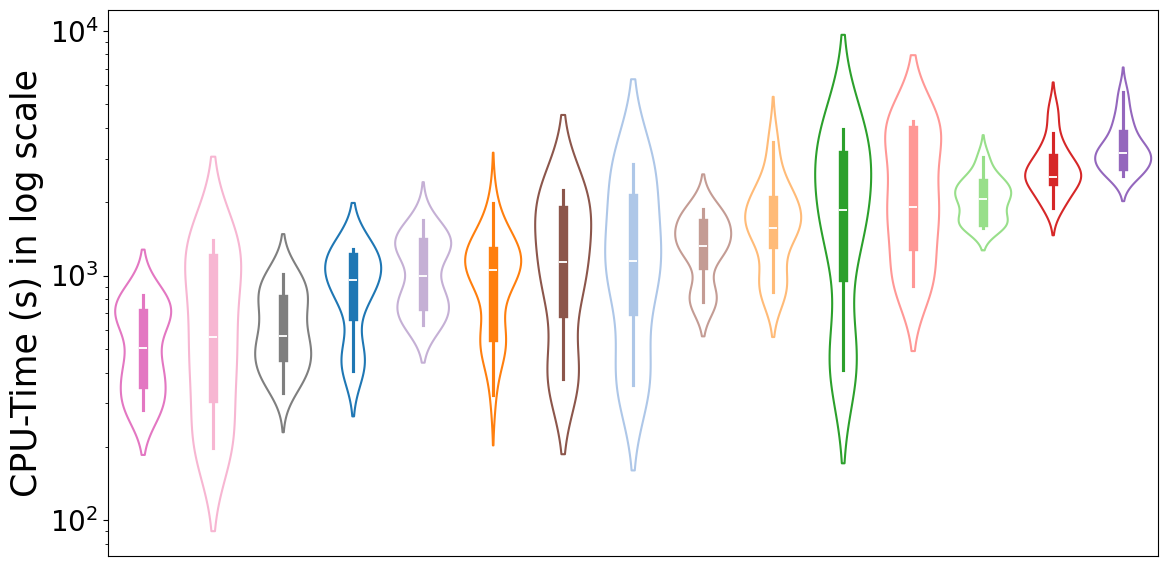}
\caption{Distribution of CPU-Time (in seconds) of the ensembles with HT, EFDT and best-ranking reported proposed method per ensemble as base learners for real-world (top) and synthetic data (down)}
\label{cc:cpu_time}
\end{figure}

Fig. \ref{cc:cpu_time} shows the distribution of CPU-Time in real-world and synthetic datasets. In real-world datasets, ARTE with HLAST$_{D}$ and ARTE with HT presented similar quantiles, while also presenting lower quantiles compared to ARF with HT and SRP with HT. The random splitting mechanism of ARTE reduces the cost of checking all possible splits, and, by simply resetting the classifier in case of concept drift detection, also lowers the cost of maintaining a background classifier. ARE with HT and HLAST$_{D}$ presented the lowest quantiles, showing the efficiency of the instance selection approach. In synthetic datasets, the difference between ARTE with HT and HLAST$_{D}$ is even more evident, as the number of instances of the synthetic datasets was $10^6$. SRP with HT presented a greater first quartile compared to ARTE with HT and HLAST$_{D}$ third quartile. ARF with HT also presented higher first quartile compared to ARTE with HT and HLAST$_{D}$ median.

\begin{figure}[h]
\centering
\begin{subfigure}[b]{1\textwidth}
          \includegraphics[width=\textwidth]{legend_1.png}
\end{subfigure} \\
\begin{subfigure}[b]{1\textwidth}
          \includegraphics[width=\textwidth]{legend_2.png}
\end{subfigure} \\
\centering
\includegraphics[width=0.8\textwidth]{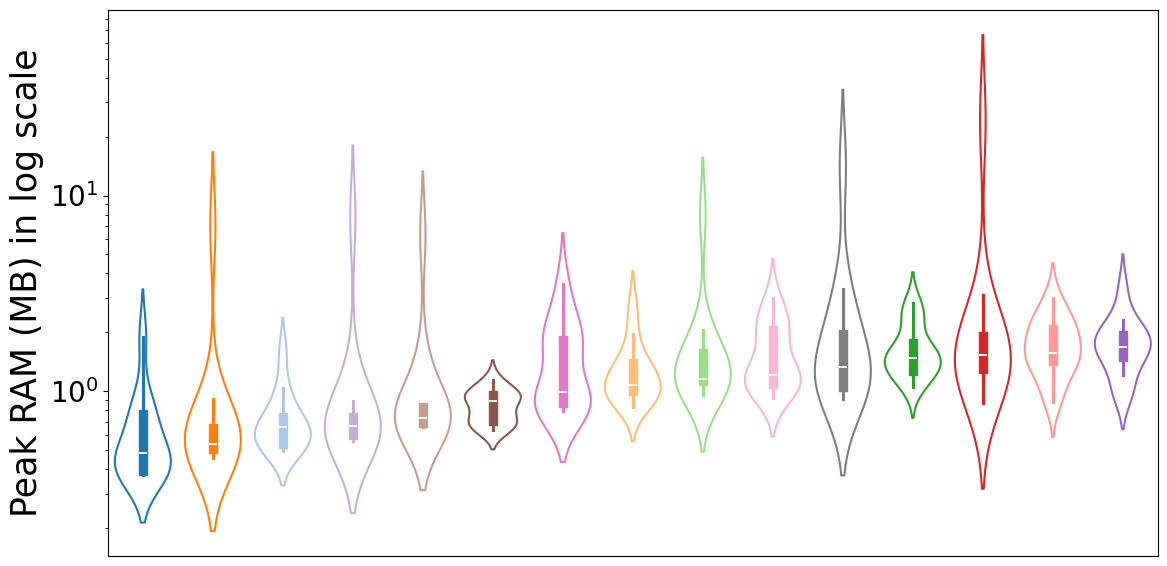}
\includegraphics[width=0.8\textwidth]{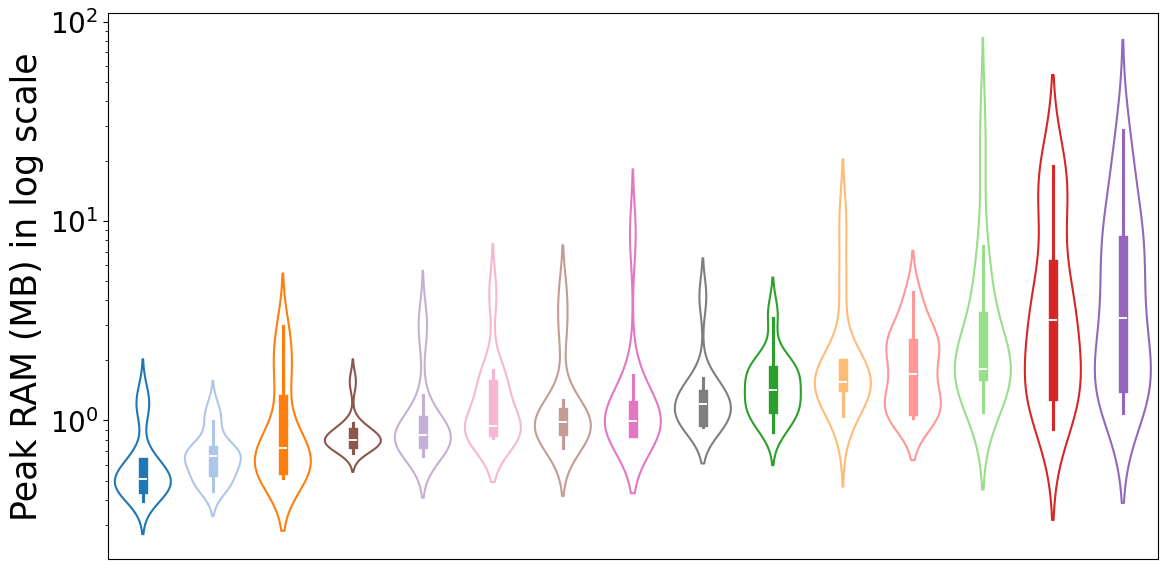}
\caption{Distribution of Peak RAM (MB) usage throughout time of the ensembles with HT, EFDT and best-ranking reported proposed method per ensemble as base learners for real-world (top) and synthetic data (down)}
\label{cc:ram_hours}
\end{figure}

Fig. \ref{cc:ram_hours} shows the distribution of peak RAM (MB) in real-world and synthetic datasets. The same pattern as for CPU-Time follows for both real and synthetic data, where ARTE with HLAST$_{D}$ and ARTE with HT present lower quantiles compared to ARF with HT and SRP with HT.

The main points outlined in this section were:
\begin{itemize}
    \item[1.] ARTE with HLAST$_{D}$ and ARTE with HT present lower quantiles compared to ARF with HT and SRP with HT for CPU-Time and RAM-Hours in real-world and synthetic data.
    \item[2.] The random splitting mechanism of ARTE reduces the cost of checking all possible splits and, by simply resetting the classifier in case of concept drift detection, also lowers the cost of maintaining a background classifier.
    \item[3.] ARE presented the lowest quantiles, showing the efficiency of the regularization approach.
\end{itemize} 

\subsection{Concept Drift}
\label{subsec:concept_drift}

\begin{figure}[H]
\centering
\begin{subfigure}[b]{1\textwidth}
          \includegraphics[width=\textwidth]{legend_1.png}
\end{subfigure} \\
\begin{subfigure}[b]{1\textwidth}
          \includegraphics[width=\textwidth]{legend_2.png}
\end{subfigure} \\
\begin{subfigure}[b]{0.30\textwidth}
\centering
          \includegraphics[width=\textwidth]{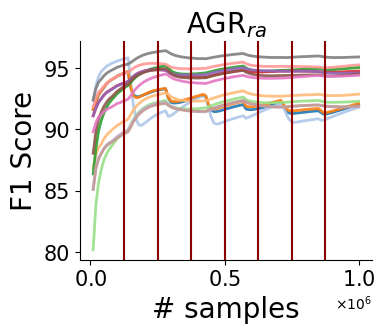}
\end{subfigure}
\begin{subfigure}[b]{0.3\textwidth}
\centering
          \includegraphics[width=\textwidth]{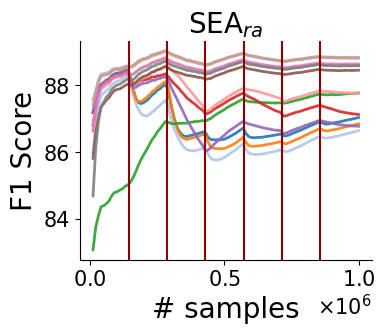}
\end{subfigure}
\begin{subfigure}[b]{0.3\textwidth}
\centering
          \includegraphics[width=\textwidth]{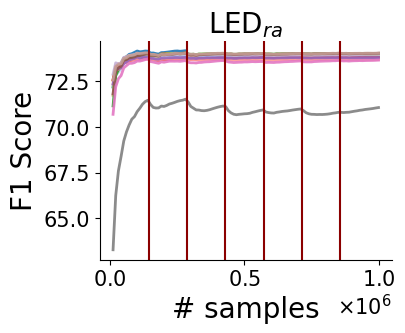}
\end{subfigure}
         \caption{F1-Score (in \%) of ensembles in datasets that simulate recurrent and abrupt changes}
   \label{drift:ens:rec_abrupt}
\end{figure}

\begin{figure}[H]
\centering
\begin{subfigure}[b]{1\textwidth}
          \includegraphics[width=\textwidth]{legend_1.png}
\end{subfigure} \\
\begin{subfigure}[b]{1\textwidth}
          \includegraphics[width=\textwidth]{legend_2.png}
\end{subfigure} \\
\begin{subfigure}[b]{0.30\textwidth}
\centering
          \includegraphics[width=\textwidth]{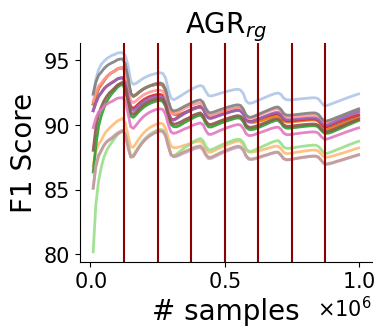}
\end{subfigure}
\begin{subfigure}[b]{0.3\textwidth}
\centering
          \includegraphics[width=\textwidth]{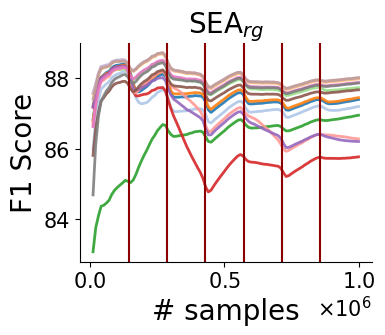}
\end{subfigure}
\begin{subfigure}[b]{0.3\textwidth}
\centering
          \includegraphics[width=\textwidth]{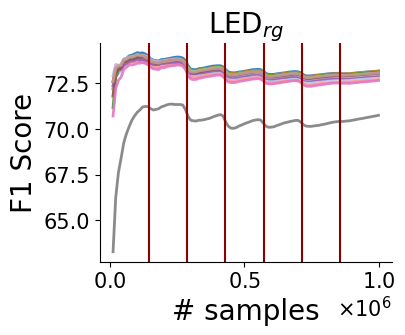}
\end{subfigure}
         \caption{F1-Score (in \%) of ensembles in datasets that simulate recurrent and gradual changes}
   \label{drift:comp:rec_gradual}
\end{figure}

\begin{figure}[H]
\centering
\begin{subfigure}[b]{1\textwidth}
          \includegraphics[width=\textwidth]{legend_1.png}
\end{subfigure} \\
\begin{subfigure}[b]{1\textwidth}
          \includegraphics[width=\textwidth]{legend_2.png}
\end{subfigure} \\
\begin{subfigure}[b]{0.3\textwidth}
\centering
          \includegraphics[width=\textwidth]{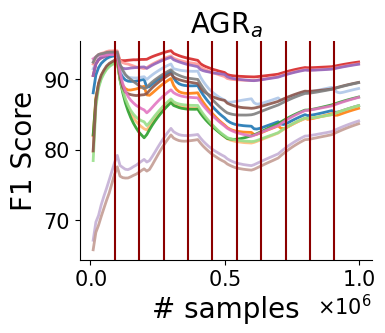}
\end{subfigure}
\begin{subfigure}[b]{0.30\textwidth}
\centering
          \includegraphics[width=\textwidth]{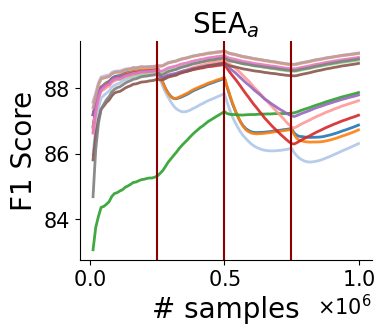}
\end{subfigure}
\begin{subfigure}[b]{0.3\textwidth}
\centering
          \includegraphics[width=\textwidth]{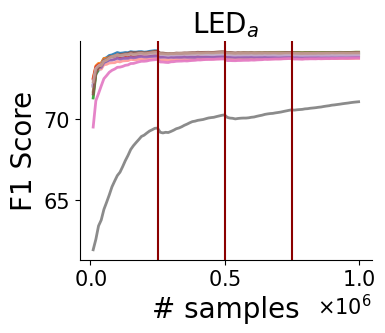}
\end{subfigure}\\
\begin{subfigure}[b]{0.3\textwidth}
\centering
          \includegraphics[width=\textwidth]{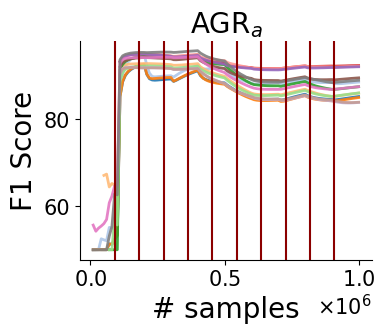}
\end{subfigure}
\begin{subfigure}[b]{0.3\textwidth}
\centering
          \includegraphics[width=\textwidth]{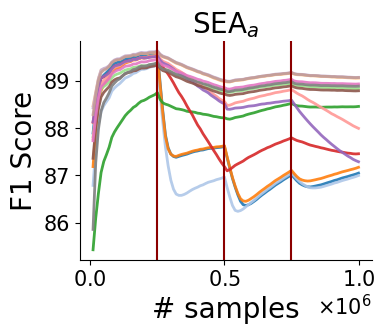}
\end{subfigure}
\begin{subfigure}[b]{0.3\textwidth}
\centering
          \includegraphics[width=\textwidth]{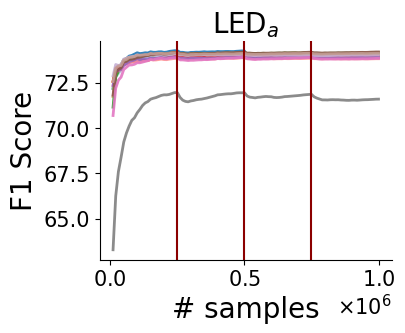}
\end{subfigure}

         \caption{F1-Score (in \%) of ensembles in datasets that simulate abrupt changes}
   \label{drift:comp:abrupt}
\end{figure}

\begin{figure}[H]
\centering
\begin{subfigure}[b]{1\textwidth}
          \includegraphics[width=\textwidth]{legend_1.png}
\end{subfigure} \\
\begin{subfigure}[b]{1\textwidth}
          \includegraphics[width=\textwidth]{legend_2.png}
\end{subfigure} \\
\begin{subfigure}[b]{0.3\textwidth}
\centering
          \includegraphics[width=\textwidth]{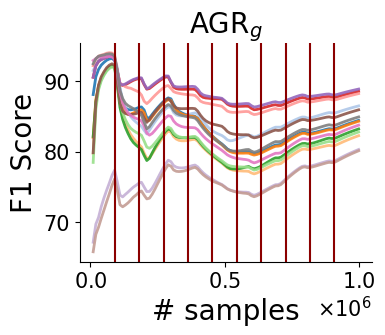}
\end{subfigure}
\begin{subfigure}[b]{0.30\textwidth}
\centering
          \includegraphics[width=\textwidth]{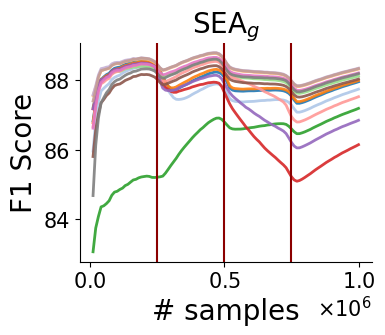}
\end{subfigure}
\begin{subfigure}[b]{0.3\textwidth}
\centering
          \includegraphics[width=\textwidth]{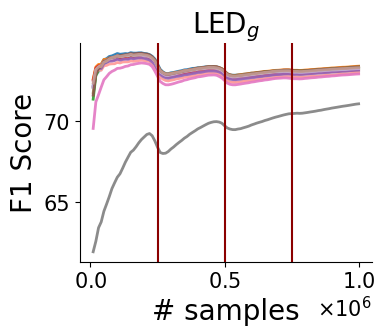}
\end{subfigure}\\
\begin{subfigure}[b]{0.3\textwidth}
\centering
          \includegraphics[width=\textwidth]{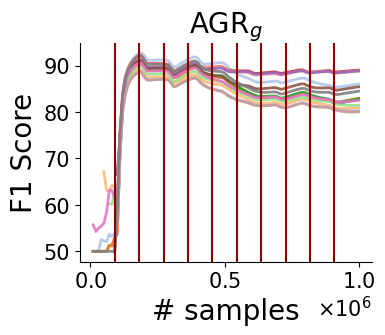}
\end{subfigure}
\begin{subfigure}[b]{0.3\textwidth}
\centering
          \includegraphics[width=\textwidth]{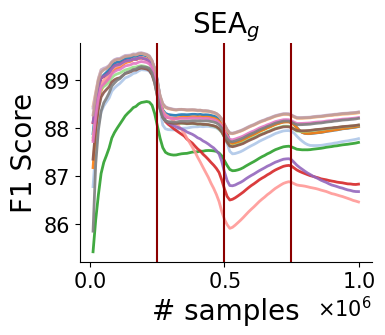}
\end{subfigure}
\begin{subfigure}[b]{0.3\textwidth}
\centering
          \includegraphics[width=\textwidth]{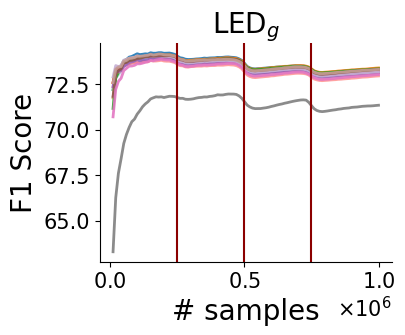}
\end{subfigure}
         \caption{F1-Score (in \%) of ensembles in datasets that simulate gradual changes}
   \label{drift:comp:gradual}
\end{figure}

Across the synthetic streams the ensembles describe very similar F1-Score curves over time, and after each drift, marked by the vertical lines in Figures \ref{drift:ens:rec_abrupt}--\ref{drift:comp:gradual}, the curves recover to a comparable level within a short window, which indicates that on these simple concepts the dominant factor is how each ensemble reacts to drift rather than which base tree it carries, in line with the small effect of the proposed trees on synthetic data reported in Section \ref{best_tree}. What separates the ensembles is how well their voting mechanism matches the structure of the current concept, and a few consistent patterns follow from the characteristics of each generator. In AGRAWAL (Fig. \ref{drift:ens:rec_abrupt}--\ref{drift:comp:gradual}) ARTE attains the lowest F1-Score, because the concepts of AGRAWAL are defined by sharp cut points on the numerical attributes and the random split-point mechanism of ARTE seldom places a split exactly on these boundaries, which is most visible in the monotone variants (Fig. \ref{drift:comp:abrupt} and \ref{drift:comp:gradual}) where the ARTE curves settle a few points below the rest. SRP behaves in the opposite way and reaches the highest F1-Score in this dataset, since the trees that were assigned the features active in the current concept fit the cut points well and, being weighted by accuracy, take over the vote.

In SEA the ordering is reversed and SRP, together with Leveraging Bagging, produces the lowest F1-Score among the ensembles right after each drift (Fig. \ref{drift:ens:rec_abrupt}, \ref{drift:comp:rec_gradual} and \ref{drift:comp:abrupt}), because the concept is defined by the sum of two relevant features and, when the drift changes which features are decisive, many trees built on the features that became irrelevant still carry a large vote weight inherited from their high accuracy under the previous concept, which delays the ensemble in tracking the new concept. In LED the curves almost overlap, with the exception of ARE, whose F1-Score stabilises a few points below the rest (the lower curve in Fig. \ref{drift:ens:rec_abrupt}), because the seventeen irrelevant features and the 10\% label noise of LED penalise the larger trees that ARE grows, which end up splitting on noisy attributes without a matching gain in leaf purity.

\begin{figure}[H]
\centering
\begin{subfigure}[b]{1\textwidth}
          \includegraphics[width=\textwidth]{legend_1.png}
\end{subfigure} \\
\begin{subfigure}[b]{1\textwidth}
          \includegraphics[width=\textwidth]{legend_2.png}
\end{subfigure} \\
\begin{subfigure}[b]{0.30\textwidth}
\centering
          \includegraphics[width=\textwidth]{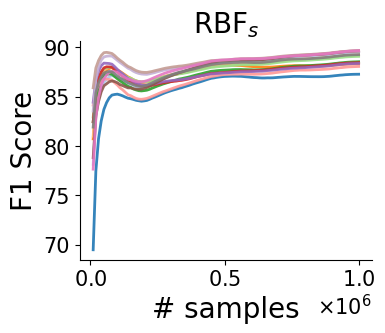}
\end{subfigure}
\begin{subfigure}[b]{0.3\textwidth}
\centering
          \includegraphics[width=\textwidth]{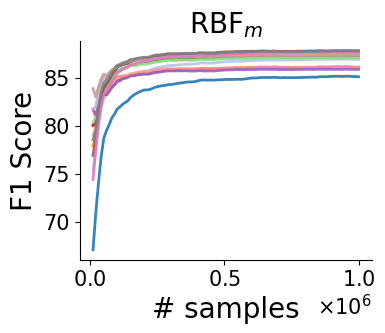}
\end{subfigure}
\begin{subfigure}[b]{0.3\textwidth}
\centering
          \includegraphics[width=\textwidth]{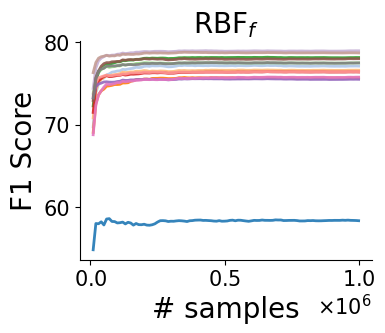}
\end{subfigure}
         \caption{F1-Score (in \%) of ensembles in the RBF dataset, which simulates incremental changes}
   \label{drift:comp:rbf_incremental}
\end{figure}

\begin{figure}[H]
\centering
\begin{subfigure}[b]{1\textwidth}
          \includegraphics[width=\textwidth]{legend_1.png}
\end{subfigure} \\
\begin{subfigure}[b]{1\textwidth}
          \includegraphics[width=\textwidth]{legend_2.png}
\end{subfigure} \\
\begin{subfigure}[b]{0.30\textwidth}
\centering
          \includegraphics[width=\textwidth]{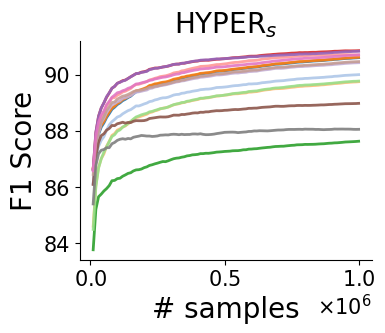}
\end{subfigure}
\begin{subfigure}[b]{0.3\textwidth}
\centering
          \includegraphics[width=\textwidth]{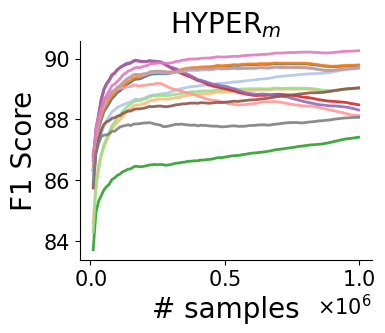}
\end{subfigure}
\begin{subfigure}[b]{0.3\textwidth}
\centering
          \includegraphics[width=\textwidth]{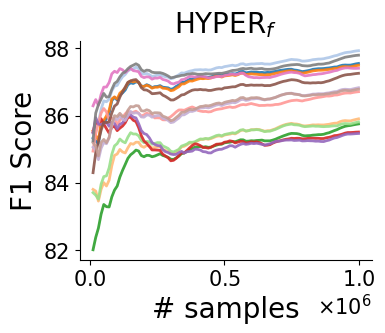}
\end{subfigure}
         \caption{F1-Score (in \%) of ensembles in the HYPER dataset, which simulates incremental changes}
   \label{drift:comp:hyper_incremental}
\end{figure}

\begin{figure}[H]
\centering
\begin{subfigure}[b]{1\textwidth}
          \includegraphics[width=\textwidth]{legend_1.png}
\end{subfigure} \\
\begin{subfigure}[b]{1\textwidth}
          \includegraphics[width=\textwidth]{legend_2.png}
\end{subfigure} \\
\begin{subfigure}[b]{0.30\textwidth}
\centering
          \includegraphics[width=\textwidth]{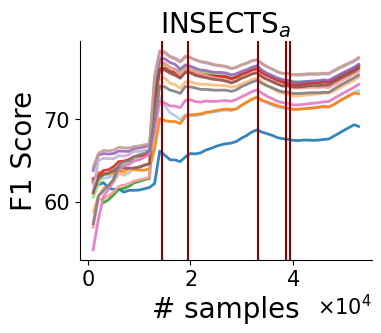}
\end{subfigure}
\begin{subfigure}[b]{0.3\textwidth}
\centering
          \includegraphics[width=\textwidth]{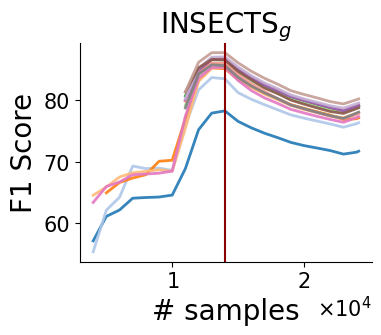}
\end{subfigure}
\begin{subfigure}[b]{0.3\textwidth}
\centering
          \includegraphics[width=\textwidth]{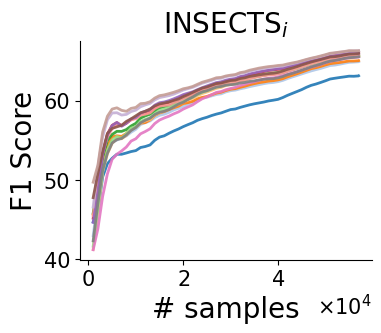}
\end{subfigure}
         \caption{F1-Score (in \%) of ensembles in the INSECTS datasets}
   \label{drift:comp:insects}
\end{figure}

On the streams with incremental drift the F1-Score curves spread more than on the abrupt ones, since the boundary moves continuously and the ensembles never settle on a fixed concept. In RBF (Fig. \ref{drift:comp:rbf_incremental}) all ensembles stay tightly grouped at a high F1-Score except Leveraging Bagging with a Hoeffding Tree base, whose curve falls further below the group as the drift speed grows, from a small gap in RBF$_s$ to a pronounced drop in RBF$_f$, where it stabilises around twenty points under the rest, while the remaining ensembles, including the proposed HLAST$_D$ variants, track the moving centroids without a comparable loss. In HYPER (Fig. \ref{drift:comp:hyper_incremental}) the spread is wider and no single ensemble dominates, but ARE is consistently among the best, because the instances misclassified during the intermediate concepts between two drifts feed additional training to its trees and the larger trees that ARE grows follow the rotating hyperplane more smoothly, whereas several ARF and ARTE configurations lag as the speed increases. In the INSECTS streams (Fig. \ref{drift:comp:insects}) the ensembles produce very close F1-Score curves and all of them climb steadily as more instances arrive, with ARTE slightly ahead.

The main findings outlined in this section were:
\begin{itemize}
    \item[1.] On the synthetic streams the ensembles describe very similar F1-Score curves that recover quickly after each drift, so the drift-reaction mechanism of the ensemble, rather than the base tree, drives the differences, and specific properties of each generator explain the gaps that do appear.
    \item[2.] ARTE struggles on datasets with sharp decision boundaries such as AGRAWAL, SRP and Leveraging Bagging lose F1-Score on the feature-dependent concepts of SEA, ARE trails on the noisy LED streams, and on the incremental drifts ARE adapts best while Leveraging Bagging with a plain Hoeffding Tree degrades sharply as the speed of the RBF drift increases.
\end{itemize}

\section{Conclusion}
\label{sec:conc}

This paper presented the Hoeffding Adaptive Splitting Trees (HLAST and EFLAST) to overcome the limitations that both Hoeffding-based trees and the adaptive LAST face as base learners of data stream ensembles. By combining the periodic split attempts of a Hoeffding Tree with an adaptive split driven by change detectors, the proposed trees keep growing at more informed moments, which promotes ensemble diversity while preserving responsiveness to concept drift.

Reporting F1-Score over a benchmark of real-world and synthetic streams across five ensembles, the experiments showed that HLAST and HLAST$_D$ are among the best-performing base trees on real-world data, surpassing HT and EFDT, while the original adaptive trees LAST and LAST$_D$ are the weakest, which confirms that pairing the adaptive mechanism with the periodic Hoeffding split is what recovers and then exceeds the performance of HT. The gains concentrate on datasets with more classes, where leaves stay impure for longer and the adaptive split has more room to act, reaching up to sixteen percentage points over HT, and on the most challenging real-world streams without temporal autocorrelation, such as Rialto, LADPU and Asfault, which shows that the improvement reflects genuine concept learning rather than artifacts of easy data. On binary and synthetic streams, whose simpler concepts are learned quickly, the proposed trees leave the results essentially unchanged.

Taken together, the proposed trees achieve the best F1-Score overall, and ARTE with HLAST$_D$ is the recommended combination, improving over ARTE with HT on almost every real-world dataset and yielding the strongest and most consistent results of the benchmark. The pairing between tree and ensemble, however, matters: HLAST$_D$, which monitors the class distribution purity, is the tree to couple with SRP, since the error-driven HLAST lets learners built on weak random feature subsets keep growing and biases the accuracy-weighted voting of SRP. This predictive quality comes at a competitive cost, as the analysis of tree size, CPU time and memory showed that ARTE with HLAST$_D$ grows larger trees than ARTE with HT but stays smaller and cheaper than SRP with HT in both CPU time and peak RAM, since its random split points avoid evaluating every candidate cut and its reset-on-drift policy spares the cost of maintaining background learners, while ARE attains the lowest cost of all through its instance-selection mechanism.

The study also provided insights into the behavior of different ensembles under various drift scenarios through their F1-Score curves over time, where the ensembles recover to comparable levels shortly after each drift and the differences that remain follow from how each approach deals with feature selection and sampling rather than from the base tree. ARTE loses F1-Score on datasets with sharp decision boundaries such as AGRAWAL, since its random split points seldom land on the well-defined cut points, while SRP and Leveraging Bagging struggle on the feature-dependent concepts of SEA, where trees built on features that became irrelevant retain a large vote weight inherited from a past concept. ARE, in turn, trails on the noisy LED streams because its larger trees split on irrelevant attributes, yet this same aggressive growth makes it the best at tracking the incremental drifts of HYPER, whereas Leveraging Bagging with a plain Hoeffding Tree degrades sharply as the speed of the RBF drift increases.

The results obtained provided insights for future work.
First, ARE has set a good direction for the proposal of new ensembles, aiming to obtain the best predictive quality with the lowest computational cost possible.
The design of extremely efficient ensembles can be further extended to:

\begin{itemize}
    \item [1.] Instead of rejecting instances, apply lower sampling. For example, misclassified instances are trained with Poisson$(\lambda = 6)$ copies, while correctly classified instances are trained with Poisson$(\lambda = 1)$ copies. The choice of varying $\lambda = 6$ and $\lambda = 1$ comes from the proposal in \cite{lev_bag}, which observed that $\lambda = 6$ provides deeper trees and higher classification accuracy, opposed to the traditional $\lambda = 1$ proposed in \cite{oza_bagging}.  
    \item [2.] Adapt ARTE with regularization techniques.
    \item [3.] Apply pre-pruning techniques to the decision tree base learners \cite{reg_ht}.
\end{itemize}

Future research directions also include adapting these methods to regression problems, tuning the hyperparameters of the base learners, and increasing ensemble diversity by selecting a different percentage of features at each node.      

\section{Acknowledgments}

This work was financed by the Pontifícia Universidade Católica
do Paraná (PUCPR) through the PIBIC Master – Combined Degree program. Finally, we sincerely thank the reviewers for their constructive feedback, which helped refine our original manuscript and strengthen the assessment of our methods.
\bibliography{sn-bibliography}

\end{document}